\documentclass[a4paper]{article}

\usepackage[UKenglish]{babel}
\usepackage[utf8]{inputenc}
\usepackage[T1]{fontenc}
\usepackage{csquotes}

\usepackage[a4paper,top=3cm,bottom=2cm,left=2cm,right=4cm,marginparwidth=3.5cm]{geometry} %

\usepackage{amsmath}
\usepackage{amsthm}
\usepackage{graphicx}
\usepackage{hyperref}

\usepackage{paralist}
\usepackage{listings}
\usepackage{cleveref}
\creflabelformat{equation}{#2#1#3}
\crefname{lstlisting}{listing}{listings}
\usepackage{booktabs}
\usepackage{cancel}
\usepackage{tikz}
\usepackage{float}

\usepackage{MarkBiblatexCmds}  %
\usepackage{MarkMathCmds}

\newcommand{\ff}{f(\cdot)}
\newcommand{\lb}{\mathfrak{L}}
\newcommand{\eye}{\mathbf{I}}
\newcommand{\K}{\ensuremath{\mathbf{K}}}
\renewcommand{\L}{\ensuremath{\mathbf{L}}}
\newcommand{\Kuu}{\ensuremath{\K_{\bf uu}}}
\newcommand{\Luu}{\ensuremath{\L_{\bf uu}}}
\newcommand{\LS}{\ensuremath{\L_{\mathbf{S}}}}
\newcommand{\kfd}[1]{\vk_{\vf#1}}
\newcommand{\kdf}[1]{\kfd{#1}\transpose}
\newcommand{\kud}[1]{\vk_{\vu#1}}
\newcommand{\kdu}[1]{\kud{#1}\transpose}
\newcommand{\kup}{\vk_{\vu f(\vx_n)}}
\newcommand{\kpu}{\kup\transpose}
\newcommand{\Kup}{\K_{\vu \mofx}}
\newcommand{\Kpu}{\Kup\transpose}

\newcommand{\Kuf}{\K_{\vu\vf}}

\newcommand{\Kff}{\K_{\vf\vf}}

\newcommand{\Kll}{\K_{\linop\linop}}

\newcommand{\Kufx}{\K_{\vu\vf}}
\newcommand{\Kfxu}{\Kufx\transpose}

\newcommand{\mS}{\mathbf{S}}
\newcommand{\mSigma}{\mathbf{\Sigma}}
\newcommand{\mofp}[1]{\boldsymbol{f}(#1)}
\newcommand{\moff}{\boldsymbol{f}}  %
\newcommand{\mof}{\moff}  %
\newcommand{\mofx}{\moff(\vx)}
\newcommand{\mofxn}{\moff({\vx_n})}

\newcommand{\mog}{\boldsymbol{g}}
\newcommand{\inspace}{\mathcal{X}}
\newcommand{\naturals}{\mathbb{N}}

\newcommand{\Mout}{{\tilde M}}

\newcommand{\mysetminusD}{\hbox{\tikz{\draw[line width=0.6pt,line cap=round] (3pt,0) -- (0,6pt);}}}
\newcommand{\mysetminusT}{\mysetminusD}
\newcommand{\mysetminusS}{\hbox{\tikz{\draw[line width=0.45pt,line cap=round] (2pt,0) -- (0,4pt);}}}
\newcommand{\mysetminusSS}{\hbox{\tikz{\draw[line width=0.4pt,line cap=round] (1.5pt,0) -- (0,3pt);}}}

\newcommand{\mysetminus}{\mathbin{\mathchoice{\mysetminusD}{\mysetminusT}{\mysetminusS}{\mysetminusSS}}}

\newcommand{\fn}[1]{f^{\mysetminus \scriptscriptstyle{#1}}\!(\cdot)}
\newcommand{\fnx}{\fn{X}}
\newcommand{\linop}{\mathcal{L}}

\newcommand{\code}[1]{\lstinline[language=Python]$#1$\xspace}
\newcommand{\GPflow}{GPflow\xspace}

\usepackage{color}

\definecolor{codegreen}{rgb}{0,0.6,0}
\definecolor{codegray}{rgb}{0.5,0.5,0.5}
\definecolor{codepurple}{rgb}{0.58,0,0.82}
\definecolor{backcolour}{rgb}{0.97,0.97,0.97}
 
\lstdefinestyle{mycodestyle}{
    basicstyle=\ttfamily\scriptsize,
    backgroundcolor=\color{backcolour},   
    commentstyle=\color{codegreen},
    numberstyle=\tiny\color{codegray},
    breakatwhitespace=false,         
    breaklines=true,                 
    captionpos=b,                    
    numbers=left,                    
    numbersep=5pt,                  
    showspaces=false,                
    showstringspaces=false,
    showtabs=false,                  
    tabsize=1
}

\title{A Framework for\\ Interdomain and Multioutput Gaussian Processes}
\author{Mark van der Wilk$^{1}$ \and Vincent Dutordoir$^2$ \and ST John$^2$ \and Artem Artemev$^2$ \and Vincent Adam$^2$ \and James Hensman$^2$}
\date{%
    $^1$Imperial College London\\%
    \texttt{m.vdwilk@imperial.ac.uk}\\[.4em]
    $^2$PROWLER.io, Cambridge, U.K.\\%
    \texttt{\{vincent, st, artem, vincent.adam, james\}@prowler.io}
}

\begin{document}
\maketitle

\begin{abstract}
One obstacle to the use of Gaussian processes (GPs) in large-scale problems, and as a component in deep learning system, is the need for bespoke derivations and implementations for small variations in the model or inference. In order to improve the utility of GPs we need a modular system that allows rapid implementation and testing, as seen in the neural network community.
We present a mathematical and software framework for scalable approximate inference in GPs, which combines interdomain approximations and multiple outputs. Our framework, implemented in GPflow, provides a unified interface for many existing multioutput models, as well as more recent convolutional structures. This simplifies the creation of deep models with GPs, and we hope that this work will encourage more interest in this approach.
\end{abstract}

\section{Introduction}
Gaussian processes (GPs) \citep{gpml} are probability distributions over functions with many properties which make them convenient for use in Bayesian models. They can be used as both priors and approximate posteriors in a wide range of models where some kind of function needs to be learned, such as classification, regression, or state-space models. While GPs are usually formulated as having a single output, multiple interacting outputs can be considered as well \citep{alvarez2012kernels}.
This is crucial for building deep Gaussian process models where we want to be able to learn more complex intermediate mappings than just simple 1D warpings.

The most well-known drawback of GPs is that exact implementations scale with the number of training points $N$ as $\BigO(N^3)$ in time. Multioutput GPs (MOGPs) have the additional problem of also needing to consider the covariances between all $P$ outputs, leading to $\BigO(N^3 P^3)$ scaling. Fortunately, the development of variational inducing variable approximations has reduced the cost of expensive matrix manipulations to $\BigO(M^3)$, where $M \ll N$, and has allowed for minibatch training \citep{titsias2009,hensman2013}. These techniques have also been extended for use in MOGPs \citep{alvarez2010} and provided large improvements. Using interdomain approximations \citep{lazaro2009inter}, a wide range of efficient inference schemes can be implemented for MOGPs, which can reduce the computational complexity to $\BigO(M^3 P)$.

In this work, we give a comprehensive introduction to sparse variational Gaussian process models, interdomain approximations, and multioutput Gaussian processes and show how they fit together in a unified description.
We then present a unified software framework for specifying interdomain approximations and multioutput models, which we have implemented in \GPflow \citep{gpflow}.
The flexible specification of interdomain approximations is particularly important for allowing the implementation of a wide range of efficient inference schemes for MOGPs, but our framework applies equally to single-output GPs.
Our immediate practical goal is to flexibly allow convolutional Gaussian process \citep{vdw2017convgp} structure in deep GPs \citep{damianou2013dgp,salimbeni2017doubly}, which requires image-to-image GPs \citep{blomqvist2018dcgp,dutordoir2019tick}. More generally, we hope that by removing obstacles to implementing and using multioutput GPs, more researchers will experiment with architectues that contain them, and that more practitioners will apply them to problems that can benefit from them.

\clearpage
\setcounter{tocdepth}{2}
\tableofcontents

\section{Scalar Gaussian processes}
Gaussian processes \citep{gpml} are most commonly discussed and used as distributions over scalar-valued functions on some input space $\inspace$, i.e.~$f: \inspace \to \Reals$. Gaussian processes are particularly convenient as a distribution over functions for two reasons:
\begin{inparaenum}[1)]
\item we can manipulate them by only considering the distribution of the function's outputs at input points of interest, and
\item the distribution of function outputs is analytically tractable, since it is Gaussian.
\end{inparaenum}
These two properties make it easy to use GPs to specify Bayesian models that involve function learning, and to develop (approximate) inference algorithms. The main advantage of specifying a distribution over functions directly, rather than specifying a distribution over parameters, is that we can use \emph{non-parametric} models, i.e.~models with an infinite number of basis functions (and corresponding parameters). This property is particularly important for obtaining uncertainty estimates which remain large in regions unconstrained by any data.

\subsection{Gaussian process notation and manipulation}
We denote a random function distributed as a GP as $f(\cdot) \sim \GP(\mu(\cdot), k(\cdot, \cdot'))$. The properties of the GP are determined by the mean function $\mu(\cdot)$ and covariance function $k(\cdot, \cdot')$. We denote an evaluation of the function $f(\cdot)$ at an input point $\vx$ as $f(\vx) \in \Reals$. Several inputs $\left\{\vx_n\right\}_{n=1}^N$ are grouped as $X$, and the evaluations at these inputs are denoted as $f(X)\in \Reals^N$. If $\inspace = \Reals^D$, then $X \in \Reals^{N\times D}$. 
A Gaussian process is defined by the property that its function values at arbitrary locations (here denoted $X$) are jointly Gaussian distributed as
\begin{align}
\p{f(X)} = \NormDist{f(X); \vmu_\vf, \Kff} \,, && \left[\vmu_\vf\right]_n = \mu(\vx_n) \,, && \left[\Kff\right]_{ij} = k(\vx_i, \vx_j) \label{eq:gp-vec-dist} \,.
\end{align}
We usually take $\mu(\cdot) = 0$ without loss of generality.

Throughout this document, we will manipulate a whole function $\ff$ as an infinitely long vector. Instead of having a finite number of elements addressed by a discrete index, we have an element for each continuous-valued input. From this view, $f(\vx)$ picks a single element out of the infinitely long vector $\ff$, while $f(X)$ picks a subset of points. We will abuse notation by denoting the Gaussian process distribution over $\ff$ as a density $p(\ff)$. Strictly speaking, it is not possible to write a density over an infinitely long vector. However, finite-dimensional marginals of the GP are well defined by densities (\cref{eq:gp-vec-dist}), so we consider $p(\ff)$ to be an object that gives us the correct finite-dimensional Gaussian distribution when we marginalise out all but a finite set of points.\footnote{Consistent marginals imply a valid measure on the whole process through the Kolmogorov extension theorem. See \citet{matthews2017thesis} for a precise discussion.} We denote this process as
\begin{align}
    p(f(X)) = \int p(\ff) \calcd\{\fnx\} \,,
\end{align}
where $\fnx$ denotes the infinitely long vector $\ff$ excluding $f(X)$. Since we work with finite datasets, we will always evaluate the GP at a finite number of locations, and therefore only need to consider a finite-dimensional marginal.

We will also often factorise $p(f(\cdot))$ into finite and infinite parts, e.g.
\begin{equation}
p(f(\cdot)) = p(\fnx \given f(X)) p(f(X)) \,.
\end{equation}
This splits the distribution over the infinitely long $\ff$ into the finite-dimensional Gaussian $p(f(X))$ (\cref{eq:gp-vec-dist}), and the infinitely long conditional $p(\fnx\given f(X))$ that includes all function values except $f(X)$. The conditional $p(\fnx\given f(X))$ is given by
\begin{align}
    p(\fnx\given f(X)) = \GP\left(\kdf{\cdot}\Kff\inv f(X)\,, \quad k(\cdot, \cdot) - \kdf{\cdot}\Kff\inv\kfd{\cdot}\right) \label{eq:gp-cond}
\end{align}
where $\left[\kfd{\cdot}\right]_n = \Cov_{f}\left[f(\vx_n), f(\cdot)\right] = k(\vx_n, \cdot)$. This expression is identical to the conditional of the whole process $p(\ff\given f(X))$. In this latter case, the random variable $f(X)$ is included in $\ff$ on the left hand side of the conditioning and has a deterministic (delta function) relationship to the conditioned value. We explicitly denote $\fnx$ in \cref{eq:gp-cond} so we do not double-count random variables on the right hand side of the equality.

\subsection{Exact inference}
\label{sec:exact-inference}
We are interested in Bayesian models that use GPs as prior distributions. Common uses are regression, classification, or unsupervised learning (e.g.~density estimation). Initially, our running example will be regression with arbitrary likelihoods.
In our setup, we model $N$ observations $\vy = \left\{y_n\right\}_{n=1}^{N}$ that are produced by evaluating the GP at input locations $\vx_n$. We denote our model as
\begin{align}
f(\cdot) \sim \GP\left(0, k(\cdot, \cdot')\right) \,, && y_n \given f(\cdot), \vx_n \sim \p{y_n \given f(\vx_n)} \label{eq:model} \,,
\end{align}
where we assume that we can evaluate the likelihood density $p(y_n\given f(\vx_n))$.

When performing inference, we are firstly interested in the posterior for making predictions, and secondly in the marginal likelihood for learning hyperparameters. One of the early reasons for interest in GPs was that closed-form solutions for these quantities exist when the likelihood is Gaussian, $p(y_n\given f(\vx_n)) = \mathcal{N}(y_n; f(\vx_n), \sigma^2)$. We can obtain the posterior for any predictive point using Bayes' rule, and the marginal likelihood through marginalising over the prior:
\begin{align}
    \p{\ff \given\vy, X} &= \frac{p(\ff)\prod_{n=1}^N p(y_n\given f(\vx_n))}{p(\vy\given X)} \nonumber \\
    &= p(\fnx\given f(X))\frac{p(f(X))\prod_{n=1}^N p(y_n\given f(\vx_n))}{p(\vy \given X)} \nonumber \\
     &= \GP\left(\kdf{\cdot}\left(\Kff + \sigma^2 \eye \right)\inv\vy, \quad k(\cdot, \cdot') - \kdf{\cdot}\left(\Kff + \sigma^2 \eye\right)\inv\kfd{\cdot'}\right) \,, \label{eq:gp-post} \\
    p(\vy\given X) &= \NormDist{\vy; 0, \Kff + \sigma^2 \eye} \label{eq:gp-marglik} \,,
\end{align}
where $\left[\kfd{\cdot}\right]_n = \Cov_{f}\left[f(\vx_n), f(\cdot)\right] = k(\vx_n, \cdot)$. To find the predictive distribution for a specific set of inputs $X^*$, we need to marginalise the posterior GP from \cref{eq:gp-post} to remove all points but $X^*$:
\begin{align}
    p(f(X^*)\given \vy, X) &= \int p(\fnx\given f(X))\frac{p(f(X))\prod_{n=1}^N p(y_n\given f(\vx_n))}{p(\vy \given X)} \calcd{\{\fn{X^*}\}} \nonumber \\
    &= \int p(f(X^*)\given f(X)) \frac{p(f(X))\prod_{n=1}^N p(y_n\given f(\vx_n))}{p(\vy\given X)} \calcd{f(X)} .
\end{align}
In \GPflow, this model is implemented as \code{GPR} (\texttt{G}aussian \texttt{P}rocess \texttt{R}egression). Note that you must include the likelihood variance when predicting actual observations, rather than the unobserved function values \citep{gpml}.  %

Despite having closed-form solutions to the quantities of interest, actual computation using these equations quickly becomes intractable for even moderately sized datasets. The cause is the $\BigO\left(N^3\right)$ computational cost for computing the inverse and determinant of $\K$.
The source of this intractability is unusual. Usually, Bayesian models have tractable priors and likelihoods, and any intractability comes from the fact that there is no closed-form solution for the normalising constant of the posterior (the marginal likelihood). The computational expense comes from needing to use methods like MCMC that avoid dealing with normalising constants. In GPs, however, it is the \emph{prior} $p(f(X))$ that is intractable, as this density already contains the costly inversion and determinant calculations.

Non-Gaussian likelihoods present an additional challenge, due to the loss of conjugacy, which allowed the closed-form computation of \cref{eq:gp-post,eq:gp-marglik}. To summarise, any approximate method for inference needs to deal with
\begin{inparaenum}[1)]
\item the cost of computing the prior, and therefore also the posterior,
\item likelihoods that are not conjugate to the Gaussian distribution.
\end{inparaenum}
We will address this in the following section.

\subsection{Approximate inference}
\label{sec:single-approx-inference}
Many approximations that address the two problems above have been proposed over the years. These can largely be categorised in terms of approximations to the model (see \citet{quinonero2005unifying} for a review), and approximations to the posterior. When approximating Gaussian process models, we need to be careful that the approximation retains the desirable properties of the original model. \Citet[][ch.~2]{vdw2019thesis} discussed desiderata that approximations to non-parametric models should satisfy, and argued that it was particularly important to maintain the uncertainty estimates afforded by using a non-parametric model. For this reason, we favour posterior approximations.

Broadly speaking, there are currently two general inference frameworks for posterior approximations to GPs that address the two problems above: expectation propagation (EP) variants \citep{minka2001expectation,csato2002sparse,bui2017unifying}, and variational inference (VI) methods \citep{blei2017variational,titsias2009,hensman2013,matthews2016sparse}. Both methods provide procedures for picking an approximate posterior from a family of tractable ones, and both methods use the same family of tractable approximate posteriors \citep{bui2017unifying}.  %
Currently, there is no consensus on which method is strictly better. In terms of pure predictive performance, neither method consistently outperforms the other. \citet{bui2017unifying} show that for approximations with few inducing points, interpolations between EP and VI often perform best, while \citet{hensman2015scalable} provide favourable evidence for VI in classification. \citet{bauer2016understanding} and \Citet{vdw2019thesis} compare EP (in its form as the FITC model approximation \citep{Snelson2006fitc,quinonero2005unifying}) and VI more closely in terms of the quality of posterior approximation. They show that EP sometimes provides a posterior approximation with significantly different properties to the true posterior. %
On the other hand, VI was shown to approximate the true posterior in a predictable way. Moreover, VI can easily be extended to importance-weighted variational inference which has been shown to converge to the true marginal likelihood \citep{Domke2018IWVI}. 
In this work, we focus on variational inference, mainly because of its practical convenience, and its strong theoretical guarantees including satisfying the desiderata from \Citet{vdw2019thesis}.

Variational inference works by minimising the KL divergence $\KL{q(f(\cdot))}{p(f(\cdot)\given \vy, X)}$ from the true posterior $p(f(\cdot)\given \vy, X)$ to a distribution in the tractable family $q(f(\cdot))$.
While the KL divergence cannot directly be computed, we can minimise it by maximising a lower bound $\lb$ to the log marginal likelihood, which has the KL as its gap:
\begin{equation}
\log \p{\vy\given X} = \lb + \KL{q(f(\cdot))}{\p{f(\cdot)\given \vy, X}} \label{eq:vi} \,.
\end{equation}
The lower bound $\lb$ depends on the variational approximation $q(\ff)$, as well as the training data and any hyperparameters. 
Minimising the KL as a means to choosing the approximate posterior provides some important properties. Firstly, since the algorithm requires only optimisation, it is guaranteed to converge. Secondly, subject to optimisation difficulties, we can only improve the approximation by expanding the family of approximating distributions \citep{bauer2016understanding,matthews2017thesis}. Finally, variational bounds are relatively easy to specify for a wide range of models, and very general algorithms exist to optimise them. These algorithms are based on stochastic evaluation of the bound \citep{hoffman2013svi} and its gradients through either score function estimators \citep{ranganath2014bbvi} or the reparameterisation trick \citep{kingma2013auto,rezende2014dglm,titsias2014doubly}.

\subsection{Inducing point posteriors}
The variational lower bound $\lb$ defined in \cref{eq:vi} depends on the family of approximate posteriors from which we will choose our approximation $q(f(\cdot))$. We need a family that contains distributions that are convenient to manipulate mathematically, as well as computationally tractable. Additionally, we would like the family to be large enough to contain a distribution very close to the true posterior. We choose the family of Gaussian process posteriors introduced by \citet{titsias2009}.

To construct a family of posteriors, we have to parameterise a distribution over functions, for which Gaussian processes are a convenient choice. We start by specifying the joint distribution over $M$ function values at the input locations $Z = \left\{\vz_m\right\}_{m=1}^M$. We collect the function values $f(Z)$ in the variable $\vu \in \Reals^M$, and specify a free mean and variance:
\begin{equation}
    q(\vu) = \NormDist{\vu; \vm, \mS} \,.
\end{equation}
Next, we specify the distribution for all other points $\fn{Z}$ using the prior conditioned on $\vu$. Together, this gives a Gaussian process on $\ff$:
\begin{align}
    q(\ff) = p(\fn{Z} \given \vu) q(\vu) = \GP\left(\kdu{\cdot}\Kuu\inv\vm, \quad k(\cdot, \cdot) - \kdu{\cdot}\Kuu\inv\left(\Kuu - \mS\right)\Kuu\inv\kud{\cdot}\right) \label{eq:q} \,,
\end{align}
where $\left[\vk_{\cdot\vu}\right]_m = k(\cdot, \vz_m)$ and $\left[\Kuu\right]_{mm'} = k\left(\vz_m, \vz_{m'}\right)$. This family of posteriors is computationally tractable, as only a smaller $M\times M$ matrix $\Kuu$ has to be inverted for making predictions.

To gain more intuition into this family of posteriors, we provide an alternative construction, similar to the unifying view from \citet{bui2017unifying}. We can consider the family of approximate posteriors from \cref{eq:q} to be \emph{the collection of all the posteriors that we can get from observing $M$ function values through an arbitrary Gaussian likelihood}. To show this, we derive \cref{eq:q} by setting up a regression problem akin to \cref{eq:gp-post} only with the likelihood $\tilde{q}(\vm'\given\vu) = \NormDist{\vm'; \vu, \mS'}$. %
We get
\begin{align}
    q(\ff) &=  p(\fn{Z}\given\vu)\frac{\tilde{q}(\vm'\given\vu)p(\vu)}{\mathcal{Z}} \label{eq:q-def-bayes} \,.
\end{align}
where $\mathcal{Z}$ is the marginal likelihood for this surrogate regression problem that normalises $q(\ff)$. Following \cref{eq:gp-post}, we obtain the approximate posterior from \cref{eq:q}
\begin{align}
    q(\ff) &= \GP\left(\kdu{\cdot}\Kuu\inv\vm', \quad k(\cdot, \cdot) - \kdu{\cdot}\left(\Kuu + \mS'\right)\inv\kud{\cdot}\right) \nonumber \\
    &= \GP\left(\kdu{\cdot}\Kuu\inv\vm', \quad k(\cdot, \cdot) - \kdu{\cdot}\Kuu\inv\left(\Kuu\inv + \Kuu\inv\mS'\Kuu\inv\right)\inv\Kuu\inv\kud{\cdot}\right) \nonumber \\
    &= \GP\left(\kdu{\cdot}\Kuu\inv\vm', \quad k(\cdot, \cdot) - \kdu{\cdot}\Kuu\inv (\Kuu - ({\mS'}\inv + \Kuu\inv)\inv)\Kuu\inv\kud{\cdot} \right) \, \label{eq:predictive},
\end{align}
where $\vm = \vm'$ and $\mS = ({\mS'}\inv + \Kuu\inv)\inv$.

This view provides a different way to intuitively understand the range of posteriors we allow. Following this interpretation, the variational method tries to find a set of $M$ observations $\vm'$ which, together with the right likelihood variance $\mS'$, give a posterior as similar as possible to the true posterior in terms of KL. This is often possible, particularly in the case where the data contain redundant information in a specific region of the input space. A concrete example of this is where we have many noisy observations in a concentrated region. In this situation, a few very precise observations can give the same posterior as many noisy observations.

We can also immediately see that when $M=N$ and a Gaussian likelihood is used in the model, the true posterior is contained within the family of distributions. \citet{burt2019} show that in this case an arbitrarily good posterior can be found as $N\to\infty$, with $M\ll N$ and the exact rate depending on the expected eigenvalue decay of $\Kff$.

It is also worth pointing out that the approximate posterior has error bars that behave in the same way as the full model. That is, both the full model, and our approximate posterior predict with an infinite number of basis functions, allowing the posterior to be uncertain in regions unconstrained by the data. This is not the case for many parametric approximations (see \citet{quinonero2005unifying} or \Citet[][ch.~2]{vdw2019thesis}).

Finally, it is worth pointing out that the choice of this family of posteriors was not accidental. This family allows the KL divergence between the $q(\ff)$ and $p(\ff\given\vy,X)$ to be finite. Many distributions over functions, including many Gaussian processes, would give an infinite KL. By choosing a family consisting of distributions that are all posteriors with respect to the prior of the model, we can guarantee a finite KL \citep{matthews2017thesis}.

\subsection{Variational inference for Gaussian processes}
\label{sec:variational-inference-for-GPs}

Now that we have specified the family of approximate posteriors, we need to find the variational lower bound. %
\Citet{matthews2016sparse} showed that, despite technically not being able to deal with densities over functions, variational inference in GP models proceeds very similarly to the well-understood parametric case. The key observation is that the KL divergence between the approximate distribution over functions (\cref{eq:q}) and the true posterior can be computed from the KL between a finite marginal distribution on function values:
\begin{equation}
    \KL{q(\ff)}{p(\ff\given\vy, X)} = \KL{q(f(X), \vu)}{p(f(X), \vu\given\vy, X)} \,.
\end{equation}
We can informally see that this makes sense, since we can include \emph{any} other set of function values (denoted $f(\cdot)$) without changing the total KL divergence:
\begin{align}
    & \KL{q(\ff, f(X), \vu)}{p(\ff, f(X), \vu\given\vy,X)} = \mathrm{KL}\big[p(\ff\given f(X), \vu)q(f(X), \vu) || \nonumber \\
    &\hspace{7.25cm} p(\ff\given f(X), \vu)p(f(X), \vu\given\vy, X)\big] \nonumber \\
    &= %
    \Exp{q(f(X), \vu)}{\underbrace{\KL{p(\ff\given f(X), \vu)}{p(\ff\given f(X), \vu)}}_{=0}} + \KL{q(f(X), \vu)}{p(f(X), \vu\given\vy, X)} \,.
\end{align}

Substituting this KL divergence into \cref{eq:vi} and applying Bayes' rule to the true posterior gives the bound discussed by \citet{hensman2013,hensman2015scalable}:
\begin{align}
    \lb = \sum_{n=1}^N \Exp{q(f(\vx_n))}{\log p(y_n \given f(\vx_n))} - \KL{q(\vu)}{p(\vu)} \label{eq:elbo-gp} \,.
\end{align}
The original variational GP method by \citet{titsias2009} is a special case for regression, where $q(\vu)$ is analytically optimised, and substituted in. The bound takes the same form as familiar bounds for parametric models, with an expected log-likelihood term and a KL to the prior. Conveniently, the final bound is no more complex than that of parametric models, despite the technical difficulties of dealing with distributions over functions.

To evaluate the bound, we need to take expectations over the approximate posterior of a single function value, given by the univariate Gaussian distribution:
\begin{equation}
q(f(\vx_n)) = \NormDist{f(\vx_n);\quad \kpu\Kuu\inv\vm,\quad k(\vx_n, \vx_n) - \kpu\Kuu\inv\left(\Kuu - \mS\right)\Kuu\inv\kup} \label{eq:qx} \,.
\end{equation}
To reduce the cost of summing over large numbers of data points, \citet{hensman2013} suggested using minibatches to obtain an unbiased estimate of the sum. Other likelihoods can be handled by estimating the expected likelihood using Monte Carlo (unbiased), or Gauss-Hermite quadrature (small bias, but zero variance) \citep{hensman2015scalable}.

\section{Interdomain Gaussian processes}
\label{sec:interdomain}
When working with Gaussian processes, we are usually concerned with observations and predictions of the function $f(\cdot)$ itself. However, point observations may not be the only properties of a function that we are interested in, or that we can observe. For example, when modelling the trajectory of an object, its velocity can be observed as well as location, which gives additional observations of the function's derivative \citep[ch.~9.4]{ohagan1992some,gpml}. Alternatively, in Bayesian quadrature, we are interested in learning about the integral of a function from point observations \citep{rasmussen2003bmc}. These quantities rely on a transformation of $\ff$ to a different domain, and we will name them ``interdomain'' variables, following \citet{lazaro2009inter}. We start by discussing how observations of these interdomain quantities can be incorporated into an exact posterior. However, we are primarily interested in using interdomain transformations to define inducing variables for specifying GP approximations, which we will discuss afterwards.
Here, we give an informal overview of interdomain methods, with a more formal discussion of some difficulties surrounding measurability being left to \citet{matthews2016sparse,matthews2017thesis}.

\subsection{Interdomain observations}
The earlier examples of derivative or integral observations can be seen
as observations of $\ff \in \Reals^\inspace$ after having been passed through some linear operator $\linop : \Reals^\inspace \to \Reals$. In other words, $\linop$ takes the entire function $\ff$ and returns a real value. We modify the likelihood from $p(y_n\given\ff,\vx_n)$ to $p(y_n\given \linop_n\ff)$, where $\linop_n$ is the linear operator for the $n$th observation.

A common example for the linear operator $\linop_n$ is an integral operator, as used by \citet{lazaro2009inter}:
\begin{align}
    \linop_n\ff = \int f(\vx) w_n(\vx) \calcd{\vx} \label{eq:id-integral} \,.
\end{align}
We can see that $\linop_n$ indeed maps the entire function $\ff$ to a single real value. A different example of a linear operator which we can express, is the derivative of the $d$th input dimension as
\begin{align}
    \linop_n\ff = \pderiv[f]{x_d}(\vx_n) \,.
\end{align}
In this case, the linear operator only depends on a small neighbourhood around $\vx_n$. Since an operator is allowed to depend on the entire function $\ff$, it is also free to only depend on a part of it while fitting within the operator framework. In fact, even regular point observations fit within this framework, where the linear operator simply depends on a single point:
\begin{align}
    \linop_n\ff = f(\vx_n) \,.
\end{align}

We will now derive the posterior for these operator observations. To ease notation, we redefine the operator $\linop$ to return all $N$ quantities needed for the likelihood terms. This makes $\linop : \Reals^\inspace \to \Reals^N$, where $\left[\linop\ff\right]_n = \linop_n \ff$. To find the posterior, we first need to find the joint distribution between variables on $\ff$, and $\linop\ff$, which the likelihood depends on. Given that the dependence between $\ff$ and $\linop\ff$ is linear, and that the prior on $\ff$ is Gaussian, the joint will be Gaussian as well. Keeping in mind the usual shorthand of $\ff$ referring to function values at arbitrary inputs, we obtain
\begin{align}
    p(\ff, \linop\ff) = p(\linop\ff\given\ff) p(\ff) = \NormDist{\begin{bmatrix}\ff \\ \linop\ff \end{bmatrix}; 0, \begin{bmatrix}k(\cdot, \cdot) & \vk_{\linop\cdot} \\ \vk_{\linop\cdot}\transpose & \Kll\end{bmatrix}} \,,
\end{align}
where the covariances $\vk_{\linop\cdot}$ and $\Kll$ are defined as
\begin{align}
    \left[\vk_{\linop\cdot}\right]_i = \Cov\left[\linop_i\ff, \ff\right] \,, && \left[\Kll\right]_{ij} = \Cov\left[\linop_i\ff, \linop_j\ff\right] \label{eq:id-covariances-def} \,.
\end{align}

These covariances can be evaluated by using the definition of $\linop\ff$, taking advantage of the linearity of $\linop$, and then taking expectations over $\ff$:
\begin{gather}
    \Cov[\linop_i\ff, f(\vx')] = \Exp{\ff}{\linop_i\ff f(\vx')} = \linop_i \Exp{\ff}{\ff f(\vx')} = \linop_i k(\cdot, \vx') \,, \label{eq:interdomain-cov} \\
    \Cov[\linop_i\ff, \linop_j\ff] = \Exp{f}{\left(\linop_i\ff\right)\left(\linop_j\ff\right)} = \linop_i k(\cdot, \cdot') \linop_j' \,, \label{eq:transdomain-cov}
\end{gather}
where with $\linop_j'$ we denote the linear operator that works on the $\cdot'$ parameter of $k(\cdot, \cdot')$.
A specific example is when all elements in $\linop$ use the integral operator of \cref{eq:id-integral}, which gives
\begin{gather}
    \Cov[\linop_i\ff, f(\vx')] \!=\! \Exp{f}{\int f(\vx)w_i(\vx)\calcd{\vx} f(\vx')} \!= \!\!\int \Exp{f}{f(\vx')f(\vx)} w_i(\vx) \calcd{\vx} \! = \!\!\int k(\vx, \vx') w_i(\vx) \calcd{\vx} \,, \\
    \Cov[\linop_i\ff, \linop_j\ff] \!=\! \Exp{f}{\!\!\left(\int\!\! f(\vx)w_i(\vx)\calcd{\vx}\right)\!\!\left(\int\!\!f(\vx')w_j(\vx')\calcd{\vx'}\right)\!\!} \!\!=\!\! \!\iint\!k(\vx, \vx') w_i(\vx)w_j(\vx') \calcd{\vx}\calcd{\vx'} .
\end{gather}

We now have all the required elements for obtaining the posterior for interdomain observations. Using the fact that $p(\ff,\linop\ff)$ is jointly Gaussian, everything reduces to Gaussian conditioning as for \cref{eq:gp-post}. We obtain the posterior\footnote{We now condition on $\linop$ to show what our observations are, analogous to conditioning on $X$ as we did earlier.}
\begin{align}
    p(\ff\given\vy,\linop) &= \int \frac{\prod_{n=1}^N p(y_n\given \linop_n\ff) p(\linop\ff\given\ff) p(\ff)}{p(\vy\given\linop)} \calcd{\{\linop\ff\}} \nonumber \\
    &= \int p(\ff\given\linop\ff)\frac{\prod_{n=1}^N p(y_n\given \linop_n\ff) p(\linop\ff) }{p(\vy\given\linop)} \calcd{\{\linop\ff\}} \nonumber \\
    &= \GP\left(\vk_{\linop\cdot}\transpose\left(\Kll + \sigma^2 \eye \right)\inv\vy, \quad k(\cdot, \cdot) - \vk_{\linop\cdot}\transpose\left(\Kll + \sigma^2 \eye\right)\inv\vk_{\linop\cdot}\right) \,, \label{eq:id-post}
\end{align}
which differs only from the point observation posterior (\cref{eq:gp-post}) through the use of the interdomain covariances $\vk_{\linop\cdot}$ and $\Kll$.

\subsection{Interdomain inducing variables}
\label{sec:interdomain-inducing}
Apart from being quantities of interest for observation or prediction, interdomain variables can also be used to specify approximate posteriors. \citet{alvarez2009sparseconvmo} used interdomain variables tailored to a particular model construction, while \citet{lazaro2009inter} discussed interdomain approximations for a given squared exponential kernel. Both these early approaches used a FITC style approach, while \citet{alvarez2010} was the first to use an interdomain posterior within the variational framework.

The construction of approximate posteriors using interdomain variables follows the exact same procedure as using inducing points. In \cref{eq:q-def-bayes} we showed that we specified our family of approximate posteriors by conditioning on $M$ arbitrary point observations. By changing the covariances as we did in \cref{eq:id-post}, we can specify our approximate posterior through conditioning on interdomain variables instead. The final form of the approximate posterior (\cref{eq:q}) is changed only through $\Kuu$ and $\vk_{\vu\cdot}$ being evaluations of the interdomain covariances of \cref{eq:id-covariances-def}. The variational lower bound remains unchanged, with the $\KL{q(\vu)}{p(\vu)}$ term now being between the interdomain variables \citep{matthews2016sparse,matthews2017thesis}.

There are many possible interdomain inducing variables that can be used to construct approximate posteriors. In certain cases, an appropriate choice can make a crucial difference to the effectiveness of an approximation, particularly when interdomain variables can be tailored to a particular kernel of interest. \citet{alvarez2010} construct a kernel through a deterministic transformation of a white noise process, and want to create an approximation by learning about the white noise process. Normal inducing points are completely ineffective at representing knowledge about white noise, but well-chosen interdomain variables can capture the useful low-frequency characteristics of a realisation. Similarly, computationally tractable convolutional GPs rely on tailored interdomain variables \citep{vdw2017convgp}. Well-chosen interdomain variables can also be used to induce computationally favourable structure in $\Kuu$ for general Matérn kernels \citep{hensman2017variational}.

In the following section we introduce multioutput GPs and discuss how interdomain approximations allow such models to scale to large numbers of data points as well as large numbers of outputs. In \cref{sec:software-framework} we then introduce our software framework that allows kernels and interdomain variables to be specified in a flexible manner, which allows all combinations to be specified with minimal code duplication.

\section{Multioutput Gaussian processes}
So far, we have discussed approximations for scalar GP models. Multioutput Gaussian processes (MOGPs) (see \citet{alvarez2012kernels} for a review) are powerful models that can represent useful correlations between related outputs. Taking these correlations into account allows the information from one output to be used in the prediction of another. This helps in multi-task learning, when predicting multiple outputs using a single model. Convolutional layers in deep convolutional Gaussian process models \citep{blomqvist2018dcgp,dutordoir2019tick} also exhibit multioutput structure, where outputs for patches in different image locations are correlated.
While the additional correlations are helpful for the model, they present an even larger computational challenge than single-output models. In this section, we will introduce MOGPs and discuss how we can do efficient inference in the variational inference framework, leading up to the discussion of our software framework in \cref{sec:software-framework}.

We are interested in models that learn a multioutput function $\mof(\cdot): \inspace \to \Reals^P$. We take the input space $\inspace$ to be $\Reals^D$, but much carries over to the general case. We denote by $f_p(\vx)$ the $p$th output of $\mof(\cdot)$ at the input $\vx$. A prior over such functions is a multioutput Gaussian process \citep{alvarez2012kernels} if the distribution of a vector of values
\begin{align}
\vf = \{f_{p_n}(\vx_n)\}_{n=1}^N\,, && \vf \in \Reals^N \label{eq:mo-collection} \,,
\end{align}
is Gaussian distributed. Note that for each element we specify the input $\vx_n$ as well as which output $p_n$ we want to consider. As with single-output GPs, multioutput GPs are fully specified by their kernel (we assume zero mean functions). Kernels for multioutput GPs can be seen as matrix-valued (or operator-valued in general) 
\begin{align}
    k:\inspace\times\inspace\to\Reals^{P\times P}\,, && k(\vx, \vx') = \Cov_{\mof(\cdot)}\left[\mofx, \mofp{\vx'}\right] \label{eq:matvalkern} \,,
\end{align}
where a covariance matrix for all outputs is returned \citep{micchelli2005learning}. This view has allowed in-depth analysis of kernel methods for learning vector- and even function-valued functions \citep{kadri2016}. An alternative view is that the matrix-valued kernel obeys the same positive definiteness properties of a single-output kernel where the index of the output is treated simply as another input. In other words, we can think of multioutput kernels as functions on the input space $\inspace$ extended by the index of the output:
\begin{align}
k: \left(\inspace, \naturals\right) \times \left(\inspace, \naturals\right) \to \Reals \,, && \Cov_{\mof(\cdot)}\left[f_p(\vx), f_{p'}(\vx')\right] = \Exp{\mof(\cdot)}{f_p(\vx) f_{p'}(\vx')} = k\left(\{\vx, p\}, \{\vx', p'\}\right) \label{eq:output-as-input} \,.
\end{align}
This means that the collection on random variables $\vf$ from \cref{eq:mo-collection} will have the distribution
\begin{align}
\p{\vf} = \NormDist{\vf; 0, \K} \,, && \mathrm{with }\quad \left[\K\right]_{ij} = k\left(\{\vx_i, p_i\}, \{\vx_j, p_j\}\right) \,.
\end{align}
This insight shows that when we observe an arbitrary collection of different output dimensions of $\mof(\cdot)$ for different inputs, which we will refer to as \emph{heterotopic} data, we can simply use the same procedure and software as for single-output GPs. 
While this view is conceptually simple, it obscures structure that can be taken advantage of for convenient and efficient implementation.

The ``output as an input'' view is particularly ill-suited for dealing with data where we observe (nearly) all output dimensions of $\mof(\cdot)$ for each input (\emph{homotopic} data), both in terms of mathematical notation and software implementation. In these situations, we will instead use
\begin{align}
    \mofp{X} = \mkvec (\{\mofxn\}_{n=1}^N ) \,, && \mofp{X} \in \Reals^{NP} \label{eq:homotopic} \,,
\end{align}
to denote the stacked outputs for all $N$ inputs $\vx_n$ in $X$. We can obtain the covariance of $p(\mofp{X})$ by taking the $(n, n')$th $P\times P$ block of $\K$ to be $k(\vx, \vx')$, using the matrix-valued kernel of \cref{eq:matvalkern}:
\begin{gather}
    p(\mofp{X}) = \NormDist{\mofp{X}; 0, \K} \,, \qquad \qquad \qquad \K \in \Reals^{NP\times NP} \,, \\
    \K = \begin{bmatrix}
    k(\vx_1, \vx_1) & \dots & k(\vx_1, \vx_n) \\ 
    k(\vx_2, \vx_1) & \dots & k(\vx_2, \vx_n) \\ 
    \vdots          & \ddots & \vdots          \\
    k(\vx_2, \vx_1) & \dots & k(\vx_2, \vx_n) \\ 
    \end{bmatrix} \label{eq:stacking} \,.
\end{gather}
We will sometimes index into $\mofp{X}$ and $\K$ as multidimensional arrays, in their unstacked form. Specifically, we denote
\begin{align}
    \left[\mofp{X}\right]_{np} = f_p(\vx_n)\,, && \left[\K\right]_{npn'p'} = k(\{\vx_n, p\}, \{\vx_{n'}, p'\}) \label{eq:indexing} \,.
\end{align}
The true index given $n$ and $p$ can be found to be $i = (n-1)P + p$.
This notation provides several benefits. Firstly, when dealing with all outputs, it is inconvenient to explicitly denote the output dimensions as inputs. Secondly, always considering all outputs may introduce structure in $\K$, which can be computationally exploited (we will see examples later). Finally, in software terms, we want to organise a collection of vector-valued outputs and their covariances in high-dimensional arrays, which makes their grouping explicit, which simplifies later use.

We will now discuss several examples of multioutput kernels. In order to take advantage of details that help efficient implementation, we take the view of evaluating a vector-valued function at particular inputs, rather than the ``output as an input'' view.
This will then lead to a software framework which
\begin{inparaenum}[a)]
\item allows easy specification of the most efficient implementation, and
\item does not compromise on unified software interfaces.
\end{inparaenum}

\subsection{Multioutput Gaussian process priors}

\citet{alvarez2012kernels} discussed different choices for priors on the correlations between multiple outputs. Here we briefly review these as well as convolutional GPs. %

\subsubsection{Linear model of coregionalization}
\label{sec:lmc}
A simple way of introducing correlations in the outputs is to construct our multioutput function $\mof(\cdot)$ from a linear transformation $W \in \Reals^{P\times L}$ of $L$ independent functions $g_\ell(\cdot)$ as:
\begin{align}
    g_\ell(\cdot) \sim \GP\left(0, k_\ell(\cdot, \cdot')\right) \,, && \mog(\vx) = \left\{g_\ell(\vx)
    \right\}_{\ell=1}^L \,, && \mofx = W\mog(\vx) \,,
\end{align}
with $\mofx \in \Reals^P$ and $\mog(\vx) \in \Reals^L$. This is known as the Linear Model of Coregionalization (LMC) \citep{journel1978mining}, and implies the covariance
\begin{align}
    k\left(\left\{\vx, p\right\}, \left\{\vx', p'\right\}\right) = \Exp{\mog}{\left[W\mog(\vx)\mog(\vx')\transpose W\transpose\right]_{pp'}} = \sum_{\ell=1}^L W_{p\ell}k_\ell(\vx, \vx') W_{p'\ell} \,.
\end{align}
This construction can be viewed as a form of input-dependent PCA or Factor Analysis, where we have a linear relationship between a latent and observed space. We can choose $L<P$ to obtain a low-dimensional explanation.

\subsubsection{Convolution processes}
\label{sec:cp}
The LMC only allows for a limited kind of correlations across outputs. It cannot capture, for example, outputs which are simply delayed versions of each other, i.e.~$f_p(\vx) = f_{p'}(\vx + \boldsymbol{\delta})$. Such more complicated relationships between outputs can still be expressed linearly by constructing $f_p(\vx)$ through a convolution of some latent process $\mog(\cdot)$. The resulting \emph{convolution process} (CP) can exhibit relationships like time-lags, and more general linear dependence on past observations. Such models have been investigated in different forms over the years \citep{higdon2002space,boyle2005dependent,alvarez2009sparseconvmo,alvarez2009lfm}. We follow the description by \citet{alvarez2010}, which constructs $\mof(\cdot)$ from a convolution of $\mog(\cdot)$ as
\begin{align}
    \mofp{\vx} = \int G(\vx - \vz) \mog(\vz) \calcd\vz \,, && \text{with } G(\vz) \in \Reals^{P\times L} \label{eq:cp} \,,
\end{align}
which, when taking the prior on $\mog(\cdot)$ as before, results in the covariance
\begin{align}
    k\left(\left\{\vx, p\right\}, \left\{\vx', p'\right\}\right) &= \Exp{\mog}{\iint G(\vx - \vz) \mog(\vz) \mog(\vz')\transpose G\left(\vx' - \vz'\right)\transpose \calcd\vz \calcd\vz'} \nonumber \\
    &= \sum_{q=1}^L \iint G_{pq}(\vx - \vz) G_{p'q}(\vx' - \vz') k_\ell(\vz, \vz') \calcd\vz \calcd\vz' \,.
\end{align}
$G(\cdot)$ is usually parameterised in a way that makes the integral tractable, and adds a number of parameters that balances flexibility against susceptibility to overfitting.

\subsubsection{Image convolutional Gaussian processes}
\label{sec:convgp}
Gaussian processes with convolutional structure \citep{vdw2017convgp} have been proposed for learning functions on images. They work by learning functions on a patch-by-patch basis, after which all patch responses are combined to form a single response for the image. This is similar to convolutional neural networks, with an exact correspondence in a particular infinite limit \citep{vdw2019thesis}.
The construction starts with a \emph{patch response function} $g(\cdot): \Reals^{w\cdot h} \to \Reals$, operating on image patches of size $w\times h$. An image response is obtained by summing the response to all $P$ patches
\begin{align}
    f(\vx) = \sum_{p=1}^P g\left(\vx^{[p]}\right) \,, \label{eq:conv-image-response}
\end{align}
where $\vx^{[p]}$ is the $p$th patch in the image $\vx$, and $P = (W\!-\!w\!+\!1)\!\times\!(\!H\!-\!h\!+\!1\!)$ is the number of patches for a $W\!\times\!H$ image. The construction can be phrased similarly to a convolution processes (\cref{sec:cp}) by taking $G(\vx, \vz) = \sum_{p=1}^P\delta\left(\vx^{[p]} - \vz\right)$, and integrating over the space of patches rather than images.

This construction is easily adapted to create an image-to-image MOGP, by simply returning all patch responses, rather than summing them. We obtain the resulting function $f(\vx): \Reals^D \to \Reals^P$ which has one output per patch. This construction was discussed by \citet{dutordoir2019tick} to create a convolutional layer for use in a deep convolutional Gaussian process model. Since we construct all outputs using the same patch response function $g(\cdot)$, the correlation of all outputs is given by the covariance function of $g(\cdot)$:
\begin{align}
    k_f\left(\left\{\vx, p\right\}, \left\{\vx', p'\right\}\right) = k_g\left(\vx^{[p]}, \vx'^{[p']}\right) \label{eq:multioutput-convgp} \,.
\end{align}

\subsection{Approximate inference}
\label{sec:multi-approx-inference}
Exact inference with MOGPs is complicated by the same issues that were discussed in \cref{sec:exact-inference}. The computational scaling in particular is worse, since if we observe $P$ outputs for all $N$ inputs, we will obtain a $PN\times PN$ covariance matrix, resulting in $\BigO(N^3P^3)$ time scaling. In the same way as for single-output GPs, these problems can be addressed in the variational framework (\cref{sec:single-approx-inference}). In the following, we will consider general likelihoods for MOGP priors of the form $p(\vy_n\given\mofxn)$, where the vector-valued observation $\vy_n$ can depend on all outputs of $\mof(\cdot)$ for the input $\vx_n$. A simple but useful example to keep in mind is where we use a \emph{correlated} Gaussian as the likelihood, e.g.:
\begin{align}
    p(\vy_n\given\mofxn) = \NormDist{\vy_n; \mofxn, \mSigma}\,, && \mSigma \in \Reals^{P\times P} \label{eq:correlated-gaussian-likelihood} \,.
\end{align}

To find the variational bound for MOGP models, we have to take this modified likelihood into account, as well as any changes that come from using multioutput variational posteriors. Luckily, much of the single-output case carries through. As before, we define the approximate posterior through the prior conditioned on the inducing observations $\vu$ (i.e.~$p(\mof(\cdot)\given\vu)$), together with a free-from density $q(\vu)$:
\begin{equation}
    q(\mof(\cdot)) = p(\mof(\cdot)\given\vu)q(\vu) \,.
\end{equation}
Denoting the prior conditonal $p(\mof(\cdot)\given\vu)$ is similar to the single-output case, but requires slightly more sophisticated notation. To obtain a similar expression, we will make the following conventions:
\begin{itemize}
    \item In multioutput scenarios, we will use $\Mout$ to refer to the number of elements in $\vu$, i.e.~$\vu \in \Reals^\Mout$. This is an important distinction for the interdomain inducing variables.
    \item We take the kernel $k(\vx, \vx')$ to be a matrix-valued kernel, as in \cref{eq:matvalkern}, i.e.~$k(\vx, \vx') \in \Reals^{P\times P}$.
    \item The covariance matrices $\Kup$ and $\K_{\vu\mofp{X}}$ are obtained by the same stacking as \cref{eq:stacking}. This makes $\Kup \in \Reals^{\Mout\times P}$, and $\K_{\vu\mofp{X}} \in \Reals^{\Mout\times NP}$.
    \item We index into these matrices in the same way as in \cref{eq:indexing}.
\end{itemize}
These conventions allow us to write the approximate posterior at $\vx$ as
\begin{align}
    \q{\mofx} = \NormDist{\mofx; \quad \Kpu\Kuu\inv\vm, \quad k(\vx, \vx) - \Kpu\Kuu\inv\left(\Kuu - \mS\right)\Kuu\inv\Kup} \label{eq:qmox} \,.
\end{align}
We can now write the multioutput variational lower bound with a variational expectation over the $P$-dimensional $\q{\mofx}$ to take the multivariate nature of the new likelihood into account:
\begin{align}
    \lb = \sum_{n=1}^N \Exp{\q{\mofxn}}{\log \p{\vy_n \given \mofxn}} - \KL{q(\vu)}{p(\vu)} \label{eq:elbo-mogp} \,.
\end{align}

This bound is written for homotopic data, as it assumes that the likelihood depends on all $P$ dimensions of $\mofx$. Heterotopic can be dealt with in the same way by simply marginalising out unobserved variables. For example, if we only observe a subset of outputs described by the index set $S_n$ at the input $\vx_n$ with the correlated Gaussian likelihood from \cref{eq:correlated-gaussian-likelihood}, we can integrate out all elements in $\vy_n$ that are not in $S_n$. The likelihood will then also only depend on a subset of the outputs in $S_n$. This only changes the variational expectation in the ELBO:
\begin{align}
    \lb = \sum_{n=1}^N \Exp{\q{\{f_s(\vx_n)\}_{s\in S_n}}}{\log \p{\left\{y_{ns}\right\}_{s\in S_n} \given \{f_s(\vx_n)\}_{s\in S_n}}} - \KL{q(\vu)}{p(\vu)} \,.
\end{align}
Integrating out the unused random variables is done by selecting the relevant rows and columns of $\Kfxu$ and $\mSigma$.

In the discussion above, we have used the vector of inducing outputs $\vu$ without defining exactly which random variables they correspond to. This general discussion only assumes that we have a covariance of $\vu$ as given by $\Kuu$. 
Creating efficient multioutput methods depends strongly on the choice of $\vu$, as this can induce structure in $\Kuu$ which can make it easier to invert. In the next section, we will discuss various choices of $\vu$ and their computational benefits.

\subsection{Example: Linear Model of Coregionalization}
\label{sec:example-lmc}
In the previous section, we stated the variational bound for multioutput GP models, and that its computational efficiency depended on the structure in $\Kuu$. Here, we discuss variants of sparse approximations to the Linear Model of Coregionalization (\cref{sec:lmc}). The main design choice in the approximation is the definition of $\vu$, which determines the matrices $\Kuu$ and $\Kuf$. We will show how different choices lead to approximations with different computational characteristics, both in terms of the ease with which $\Kuu$ can be inverted, as well as how many kernel elements need to be computed. This section discusses examples in mathematical terms, while \cref{sec:lmc-implementation} delves into the implementation.

\subsubsection{Inducing points}
\label{sec:example-lmc-inducing-points}
We start with by introducing two formulations of inducing points for MOGPs. In the same way as in single-output GPs, these inducing points are general, and work with every multioutput kernel.

We can choose our inducing variables in an unstructured way, by specifying $\Mout$ input-output pairs $\{\vz_m, p_m\}$ using the ``output as an input'' formulation. This gives the following $\vu$ and corresponding covariance matrices:
\begin{align}
    \vu = \{f_{p_m}(\vz_m)\}_{m=1}^\Mout \,, && \left[\Kuu\right]_{mm'} = k(\vz_m, \vz_{m'})_{p_mp_{m'}} \,, && \left[\Kup\right]_{mp} = k(\vz_m, \vx)_{p_mp} \,.
\end{align}
Alternatively, we can define inducing points for MOGPs as specifying $M$ inducing inputs, and using \emph{all} corresponding outputs as inducing variables:
\begin{align}
     \vu = \mofp{Z} \,, &&
     \left[\Kuu\right]_{mp,m'p'} = k(\vz_m, \vz_{m'})_{pp'} \,, &&
     \left[\Kup\right]_{mp,p'} = k(\vz_m, \vx)_{pp'} \,.
\end{align}
In this case, we have $\Mout = MP$ inducing variables for $M$ inducing inputs. This ``vector-valued inducing point'' approach defines more inducing variables for fewer inducing inputs, which reduces the number of parameters to be optimised. This may be beneficial even when dealing with heterotopic data, particularly if the input distribution is the same for all outputs.

Both these methods yield dense $\Kuu$ and $\Kup$ matrices for which no additional simplifying assumptions can be made. However, this method does immediately work for all multioutput kernels.
We include the vector-valued inducing point approach in GPflow (\cref{sec:lmc-implementation-inducingpoints}) as the most basic and general method.

\subsubsection{Latent inducing points}
\label{sec:example-lmc-latent-inducing-points}
For the LMC, approximate inference can be made more efficient by a more judicious choice of inducing variables. In \cref{sec:lmc}, we constructed the LMC from $L$ \emph{independent} latent processes $\mog(\cdot)$. If we choose evaluations of $\mog(\cdot)$ as our inducing variables, we can take advantage of the prior independence of $g_\ell(\cdot)$ over $\ell$. We collect all outputs of $\mog(\cdot)$ for $M$ inputs in Z, to obtain $\Mout = ML$ inducing variables. Thanks to the prior independence, $\Kuu$ will have a block diagonal structure, meaning that the largest matrix that has to be inverted only has size $M \times M$. We specify the approximation as
\begin{align}
    \vu = \mog(Z) \,, && \left[\Kuu\right]_{m\ell,m'\ell'} = k_\ell(\vz_m, \vz_{m'})\delta_{\ell\ell'} \,, && \left[\Kup\right]_{m\ell,p} = W_{p\ell}k_\ell(\vz_m, \vx) \,.
\end{align}
We impose an additional mean-field restriction on $q(\vu)$ by choosing $\mS = \blockdiag_{\ell=1}^L\left[\mS_\ell\right]$, where $\mS_\ell \in \Reals^{M\times M}$. This gains extra computational efficiency at the cost of some accuracy. We can now rewrite \cref{eq:qmox} to take advantage of this block-diagonal structure. We write ${\Kuu\inv}_\ell$ for the inverse of the $\ell$th ${M}\times M$ block of $\Kuu$. %
\begin{align}
    q(\mofx) &= \NormDist{\mofx; \vmu, \mSigma} \,, \\
    \left[\vmu\right]_p &= W_{p\ell} k_\ell(\vz_m, \vx) \left[{\Kuu\inv}_\ell\right]_{mm'} \left[{\vm}\right]_{m'\ell} \,, \label{eq:lmc-latent-ip-mu} \\
    \left[\mSigma\right]_{pp'} &= k(\vx, \vx)_{pp'} - W_{p\ell}k_\ell(\vz_m, \vx)\left[{\Kuu\inv}_\ell\left({\Kuu}_\ell - \mS_\ell\right){\Kuu\inv}_\ell\right]_{mm'} k_\ell(\vz_{m'}, \vx)W_{p'\ell} \label{eq:lmc-latent-ip-cov} \,,
\end{align}
where we implicitly sum over repeated indices (Einstein notation). Furthermore, by noting that $k(\vx, \vx)_{pp'} = W_{p\ell}k_\ell(\vx, \vx)W_{p'\ell}$, we can write \cref{eq:lmc-latent-ip-mu,eq:lmc-latent-ip-cov} as a linear transformation of the predictive distributions of $\mog(\cdot)$. If we denote the predictive density at the input $\vx$ as $q(\mog(\vx)) = \NormDist{\mog(\vx), \vmu_\vg, \mSigma_\vg}$ (note that $\mSigma_\vg$ is diagonal), we have
\begin{align}
    \vmu = W\vmu_\vg \,, && \mSigma = W\mSigma_\vg W\transpose \label{eq:transformation-g} \,.
\end{align}

\subsubsection{Latent inducing points for the Intrinsic Model of Coregionalisation}
\label{sec:example-lmc-latent-inducing-points-shared}
We can further gain computational benefits by constraining the kernels of each $g_\ell(\cdot)$ to be equal, which is known as the \emph{Intrinsic Model of Coregionalisation} (IMC). In terms of modelling quality, this constraint may be beneficial or a hindrance depending on the characteristics of the dataset. However, it is much faster than the more flexible model of the previous section: instead of $L$ separate inversions, only a single $M\times M$ inversion has to be performed. Our predictive mean and covariance become
\begin{align}
    \left[\vmu\right]_p &= W_{p\ell} k(\vz_m, \vx) \left[{\Kuu\inv}\right]_{mm'} \left[{\vm}\right]_{m'\ell} \,, \\
    \left[\mSigma\right]_{pp'} &= k(\vx, \vx)_{pp'} - W_{p\ell}k(\vz_m, \vx)\left[\Kuu\inv\left(\Kuu - \mS_\ell\right)\Kuu\inv\right]_{mm'} k(\vz_{m'}, \vx)W_{p'\ell} \,.
\end{align}
We can again rewrite these terms as in \cref{eq:transformation-g}, only in this case the computation of $\vmu_\vg$ and $\mSigma_\vg$ would require only a single inversion.

\subsubsection{Computational complexity}
\label{sec:example-lmc-computational-complexity}
The choice of inducing variable has a large impact on the computational complexity of the method. We compare the methods of the previous three sections in \cref{tab:lmc-computational-complexity}. For the methods in \cref{sec:example-lmc-latent-inducing-points,sec:example-lmc-latent-inducing-points-shared}, we have two ways to compute the required quantities:
\begin{inparaenum}[1)]
\item first calculate $\K_{\vu\mofp{X}}$ followed by a summation over $mm'\ell p$,
\item first sum over $mm'$ to obtain $\vmu_\vg$ and $\mSigma_\vg$, and then sum over $\ell$.
\end{inparaenum} While method 2 is more efficient than method 1, this choice influences the complexity of implementation, which we discuss in \cref{sec:lmc-implementation}.

\begin{table}[htb]
\centering
\begin{tabular}{llll}
\hline
      Method & Vector IP (\S\ref{sec:example-lmc-inducing-points}) & Latent IP (\S\ref{sec:example-lmc-latent-inducing-points}) & Latent IP -- ICM (\S\ref{sec:example-lmc-latent-inducing-points-shared})  \\
\hline
1)    & $\BigO\left(P^3M^3 + NP^3M^2\right)$  & $\BigO\left(LM^3 + NPLM^2\right)$ & $\BigO\left(M^3 + NPLM^2\right)$ \\
2)         & --  & $\BigO\left(LM^3 + NLM^2 + NPL\right)$   & $\BigO\left(M^3 + NLM^2 + NPL\right)$        \\
\hline
\end{tabular}
    \caption{Computational complexity of computing $\vmu$ for different choices of inducing points (IP) for the LMC model. We explicitly include the dependence on the number of predictions $N$ that need to be made.}
    \label{tab:lmc-computational-complexity}
\end{table}

\section{Software framework}
\label{sec:software-framework}

In this section, we will present our unified implementation of interdomain and multioutput Gaussian processes, following the mathematical derivation that we described so far. Our framework is implemented as a large extension to \GPflow \citep{gpflow}.
We start with describing the desiderata which guided our design choices, followed by a description of the code architecture. We will specifically refer to locations in the code where key items are implemented, so the reader can gain an idea for how other kernels and interdomain variables can be implemented.

\subsection{Desiderata}
Our contribution to \GPflow maintains its limited scope of implementing only variational approximations to GP models. While many of the abstractions described here are useful for other methods (e.g.~EP), we maintain the narrow focus to ensure high-quality code. Even with this limitation, the previous sections illustrated that a wide variety of models can be approximated with a wide variety of approximate posteriors. The goal of our software framework is to facilitate the implementation of as many combinations of model and approximation structure as possible, in a computationally efficient manner. We summarise this goal using three desiderata:
\begin{enumerate}[1)]
\item Efficiency: Our framework must allow the most computationally efficient code to be implemented, by exploiting any mathematical structure in the model and approximation.
\item Modularity: Using different combinations of modelling and approximation assumptions should require a minimum of effort and code-duplication.
\item Extensibility: Our framework should allow new model and approximation variations to be implemented without modifying core \GPflow source, following the open/closed design principle \citep{martin1996open}.
\end{enumerate}

The mathematical framework discussed in the previous sections requires us to make three choices that influence the overall method. The kernel and likelihood determine model properties, while the choice of interdomain variables $\vu$ determines properties of the approximation.
We will achieve modularity and extensibility by expressing the design choices as class hierarchies, which specify the most efficient code path through \emph{multiple dispatch}.

\subsection{\GPflow model setup and examples}
\label{sec:gen-structure}

We focus our implementation around the variational bound from  \cref{eq:elbo-gp,eq:elbo-mogp} \citep{hensman2013,hensman2015scalable}, as it works for general kernels, inducing variables, and likelihoods, in contrast to more specialised methods like \citet{titsias2009}.
Our contribution extends the existing \texttt{SVGP} (\texttt{S}parse \texttt{V}ariational \texttt{GP}) implementation in GPflow \citep{gpflow}. \texttt{SVGP} requires that three main components are specified:
\begin{enumerate}
    \item the kernel, which determines the prior on $\mof(\cdot)$,
    \item the inducing variables $\vu$, and
    \item the likelihood.
\end{enumerate}
Each is passed as an object to the \code{SVGP}, presenting a flexible interface for users to combine different kernels, inducing variables, and likelihoods. A model can be trained to the dataset \texttt{(X, Y)} by minimising its objective function with respect to its variational and hyperparameters, using:
\begin{lstlisting}[language=Python, style=mycodestyle]
optimizer = tf.optimizers.Adam()

@tf.function
def loss_closure():
    return - model.elbo((X, Y))
    
optimizer.minimize(loss_closure, model.trainable_variables)
\end{lstlisting}
We now review several examples that show how our interface can be used, and that it allows for the creation of models that flexibly mix and match different components.

\paragraph{Single-output and inducing points}
\label{sec:api-single-output}
The most straightforward use case is simple single-output regression with inducing points on a dataset where \code{X.shape == (N, D)} and \code{Y.shape == (N, 1)}. This model can be set up with \cref{lst:api-single-output}.

\begin{lstlisting}[language=Python, style=mycodestyle, mathescape,label=lst:api-single-output,caption={Set-up code for a single-output GP model, with an inducing point posterior.}]
kernel = gpflow.kernels.SquaredExponential()
Z = X[:M, :].copy()  # Initialise inducing locations to the first M inputs in the dataset
inducing_variable = gpflow.inducing_variables.InducingPoints(Z)
likelihood = gpflow.likelihoods.Gaussian()
model = gpflow.models.SVGP(kernel, likelihood, inducing_variable)
\end{lstlisting}

\paragraph{Single-output and Fourier features}
The properties of the inducing variables greatly influence the computational properties of the method. Variational Fourier Features \citep{hensman2017variational}, for example, reduce the cost of inverting $\Kuu$, and allow precomputation of many costly quantities. We can use them by simply changing the inducing variable that is passed into \texttt{SVGP} in \cref{lst:api-single-output}:
\begin{lstlisting}[language=Python, style=mycodestyle]
inducing_variable = FourierFeature1D(num_freqs=(M+1)//2)
\end{lstlisting}
with everything else remaining the same. We give the complete example in the \citet{gpflow-vff-notebook}.

\paragraph{Convolutions and inducing patches} We can create a GP model with convolutional structure \citep{vdw2017convgp} by using a \code{Convolutional} kernel:
\begin{lstlisting}[language=Python, style=mycodestyle]
kernel = gpflow.kernels.Convolutional(
    gpflow.kernels.SquaredExponential(),
    img_size=(H, W), 
    patch_size=(h, w)
)
\end{lstlisting}
where a set of $N$ images is represented as a matrix \code{X} of shape $N \times (W \cdot H)$. While using ``default'' \code{InducingPoints} is possible, \citet{vdw2017convgp} argue that tailoring the inducing variables to the convolutional structure of the kernel leads to a much more efficient method. 
We achieve this by instead using \texttt{InducingPatches} as inducing variables:
\begin{lstlisting}[language=Python, style=mycodestyle]
Z = extract_random_patches_from_images(X)
inducing_variable = InducingPatches(Z)
\end{lstlisting}
A fully-working example can be found in the \citet{gpflow-conv-notebook}.

\paragraph{Multioutput kernels}
Multioutput models also fit into this framework, and can be specified using a multioutput kernel. The usual \texttt{InducingPoints} can still be used to specify a posterior, as described in \cref{sec:example-lmc-inducing-points}, %
resulting in the code:
\begin{lstlisting}[language=Python, style=mycodestyle]
kernel_list = [
    gpflow.kernels.SquaredExponential(D) for _ in range(L)
]  # specify a kernel for each output
W = np.random.randn(P, L)  # Linear mixing of the GP outputs
kernel = gpflow.kernels.LinearCoregionalisation(kernel_list, W=W)
iv = gpflow.inducing_variables.InducingPoints(Z)
\end{lstlisting}
Alternatively, we can use inducing points defined in $\mog(\cdot)$ to take advantage of the most efficient code path (\cref{sec:example-lmc-latent-inducing-points}):
\begin{lstlisting}[language=Python, style=mycodestyle]
from gpflow.inducing_variables import InducingPoints, SharedIndependentInducingVariables
iv = SharedIndependentInducingVariables(InducingPoints(Z))
\end{lstlisting} 
See the \citet{gpflow-multioutput-notebook} for more examples on multi-output GPs in GPflow.

\subsection{Implementation of \texttt{SVGP}}
\label{sec:implementation-svgp}

\begin{figure}
    \centering
    \includegraphics{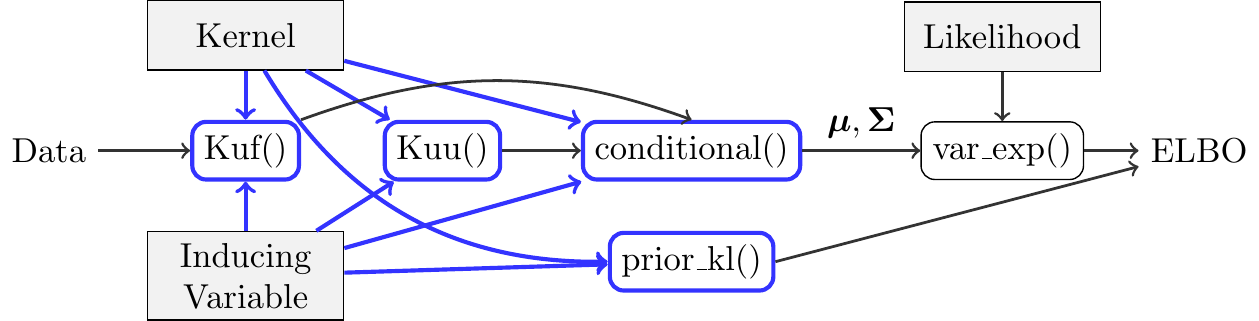}
    \caption{Flow of \code{elbo()}.}
    \label{fig:elbo-flow}
\end{figure}

The main role of \code{SVGP} is to specify the computation of the training objective, and to allow for future predictions once the model is trained. As stated earlier, we train our model by maximising the ELBO (\cref{eq:elbo-gp,eq:elbo-mogp}), which we reproduce here:
\begin{equation}
\lb = \sum_{n=1}^N \Exp{q(\mofxn)}{\log p(\vy_{n}\given \mofxn)} - \KL{q(\vu)}{p(\vu)}\,. \label{eq:soft:elbo}
\end{equation}
We need to implement \code{SVGP} and its components in a way that satisfies our desiderata of efficiency, modularity, and extensibility. The example use-cases in the previous section determine the interface of \code{SVGP} to ensure modularity. To satisfy extensibility, we need to ensure that \code{SVGP} does not rely on properties that are specific to a particular kernel or inducing variable, and that new behaviour can be implemented without modifications to \GPflow. We will separately discuss each of the four components that form the ELBO:
\begin{inparaenum}[1)]
\item the KL term,
\item the predictive distribution $\q{\mofx}$,
\item the expectation of the log-likelihood over the predictive distribution, and
\item the sum of the expectations over the log-likelihood.
\end{inparaenum} Although slightly hidden from the implementation of \code{SVGP}, we will also discuss how the covariance matrices $\Kuu$ and $\Kuf$ are computed, which the KL term and predictive distribution rely on.

We see a skeleton of the implementation of \code{SVGP} in \cref{lst:svgp}. The interface of the constructor permits the usage described in the previous section, by accepting a kernel, inducing variable, and likelihood. Optionally, a user can specify a mean function, and a different parameterisation of the variational parameters (using \code{whiten}, see \cref{sec:software:kl}). The core of \code{SVGP} computes the objective function in \code{SVGP.elbo()}, where we see the separation of the different terms of the ELBO:

\begin{itemize}
\item \textbf{KL divergence} The first quantity we compute in \code{elbo()} is the $\KL{q(\vu)}{p(\vu)}$ term (\cref{eq:soft:elbo}), i.e.~the KL divergence between the approximate posterior and prior of the inducing outputs. This can be calculated in closed form as both distributions are Gaussian. The method that computes this quantity, \code{kullback_leiblers.prior_kl()} takes the inducing variable and kernel objects, in addition to the specification of the variational posterior through \code{q_mu} and \code{q_sqrt}. We pass the inducing variable and kernel in as objects so we can specialise the computational path to take advantage of any resulting structure in $\Kuu$.

\item \textbf{Predictive distribution} The predictive distribution $\q{\mofx}$ is computed through a call to the function \code{conditional()} inside \code{predict_f()}. The method will return the mean \code{fmean} and the variance \code{fvar} of the posterior predictive GP evaluated at points \code{Xnew}, following the general equation of a conditional GP \cref{eq:predictive}. As with the KL divergence, we want to tailor the code path to the combination of inducing variable and kernel that is being used, so we pass these objects in to the function as parameters.

\item \textbf{Expected log-likelihood} The expected log-likelihood is encapsulated in classes inheriting from \code{Likelihood}, which only requires the predictive distribution (\texttt{fmean} and \texttt{fvar}), and is therefore independent of any model choices once \code{conditional()} has been called.

\item \textbf{Minibatch sum}
In the last line of \code{elbo()} we compute the sum of the expected log-likelihoods of the data points in the batch and rescale its contribution to the overall ELBO to compensate for the minibatching. 
\end{itemize}

\begin{lstlisting}[language=Python, style=mycodestyle, caption={\GPflow implementation of \texttt{SVGP}.},label=lst:svgp]
class SVGP(GPModel):
    def __init__(self, kernel, likelihood, inducing_variables=None, mean_function=None, whiten=True, ...):
        ... # Initialise

    def prior_kl(self) -> tf.Tensor:
        return kullback_leiblers.prior_kl(self.inducing_variables, self.kernel,
                                          self.q_mu, self.q_sqrt,
                                          self.whiten)

    def maximum_likelihood_objective(self, data: Tuple[tf.Tensor, tf.Tensor]) -> tf.Tensor:
        return self.elbo(data)

    def elbo(self, data: Tuple[tf.Tensor, tf.Tensor]) -> tf.Tensor:
        X, Y = data
        kl = self.prior_kl()
        fmean, fvar = self.predict_f(X, full_cov=False, full_output_cov=False)
        var_exp = self.likelihood.variational_expectations(fmean, fvar, Y)
        scale = ...  # self.num_data divided by minibatch size
        return tf.reduce_sum(var_exp) * scale - kl

    def predict_f(self, Xnew: tf.Tensor, full_cov=False,
                  full_output_cov=False) -> Tuple[tf.Tensor, tf.Tensor]:
        q_mu = self.q_mu
        q_sqrt = self.q_sqrt
        mu, var = conditional(Xnew, self.inducing_variables, self.kernel,
                              q_mu, q_sqrt=q_sqrt,
                              full_cov=full_cov, full_output_cov=full_output_cov,
                              white=self.whiten)
        return mu + self.mean_function(Xnew), var
\end{lstlisting}

We will now delve into the implementation of these terms in more detail.

\subsubsection{Multiple dispatch}
\label{sec:multipedispatch}
Both for computing the KL divergence (\code{prior_kl()}) and for computing the predictive density (\code{conditional()}), the code path depends on the type of inducing variables and kernel we use. One possible approach for customising the code paths inside these functions would be to have branches for each pair that is implemented, e.g.:
\begin{lstlisting}[language=Python, style=mycodestyle]
def prior_kl(inducing_variables, kernel, q_mu, q_sqrt, whiten=False):
    if isinstance(inducing_variables, gpflow.inducing_variables.FourierFeature1D) and isinstance(kernel, gpflow.kernels.Matern12):
        # Take advantage of low-rank + diagonal structure in Kuu
        ...
    elif isinstance(inducing_variables, gpflow.inducing_variables.SharedIndependentInducingVariables) and isinstance(kernel, gpflow.kernels.LinearCoregionalization):
        # Take advantage of block-diagonal structure in Kuu, etc...
        ...
    elif ...:
        ...
    else:
        raise NotImplementedError
\end{lstlisting}
While this implementation would allow the most efficient code paths to be specified, it is not extensible. We want to be able to implement specific code for new combinations of kernel and inducing variable without having to edit core GPflow code.

To provide this functionality, we rely on \emph{multiple dispatch} (we use the Python implementation by \citet{multipledispatch}). When a multiple dispatch function is called, the runtime types of its arguments are used to determine which specific code path to run. This is a runtime equivalent of function overloading in statically typed programming languages. Multiple dispatch allows new code paths to be registered as separate functions, which can be placed anywhere in code, therefore enabling extensibility. A code path for a new pair of types is registered using the \code{dispatcher.register(*types)} decorator. For example, in GPflow we register various implementations of \code{conditional()}:
\begin{lstlisting}[language=Python, style=mycodestyle,caption={Examples of several tailored code paths of \texttt{conditional}.},captionpos=b,label=lst:multidispatch-conditional-example]
@conditional.register(object, InducingVariables, Kernel, object)
def singleoutput_conditional(Xnew: tf.Tensor, inducing_variable: InducingVariables, kernel: Kernel, f: tf.Tensor, *, full_cov=False, full_output_cov=False, q_sqrt=None, white=False):
    """Single-output GP conditional.

    The covariance matrices used to calculate the conditional have the following shape:
    - Kuu: [M, M]
    - Kuf: [M, N]
    - Kff: [N, N]"""
    ...

@conditional.register(object, InducingPoints, MultioutputKernel, object)
def inducing_point_conditional(Xnew, inducing_variable, kernel, f, *, full_cov=False, full_output_cov=False, q_sqrt=None, white=False):
    """Multi-output GP with fully correlated inducing variables.
    The inducing variables are shaped in the same way as evaluations of K, to allow a default
    inducing point scheme for multi-output kernels.
    The covariance matrices used to calculate the conditional have the following shape:
    - Kuu: [M, L, M, L]
    - Kuf: [M, L, N, P]
    - Kff: [N, P, N, P], [N, P, P], [N, P]"""
    ...
    
@conditional.register(object, SharedIndependentInducingVariables, LinearCoregionalization, object)
def coregionalization_conditional(Xnew, inducing_variable, kernel, f, *, full_cov=False, full_output_cov=False, q_sqrt=None, white=False):
    """Most efficient routine to project L independent latent gps through a mixing matrix W.
    The mixing matrix is a member of the `LinearCoregionalization` and has shape [P, L].
    The covariance matrices used to calculate the conditional have the following shape:
    - Kuu: [L, M, M]
    - Kuf: [L, M, N]
    - Kff: [L, N] or [L, N, N]"""
    ...
\end{lstlisting}
When \code{conditional()} is called, the appropriately typed implementation from above is run. The inheritance structure of the inducing variable and kernel classes is respected, with the most specific type that has an implementation being preferred, and parent types providing default implementations. For example, if a new inducing variable \code{NewInducingVariable} inherits from \code{InducingPoints}, \code{conditional()} will call the second implementation in \cref{lst:multidispatch-conditional-example}, unless a more specific implementation is registered with \texttt{@conditional.register(NewInducing\allowbreak{}Variable,\allowbreak{} kernels.\allowbreak{}Kernel)}.

The key benefits of implementing \code{SVGP} in this way, is that the model is completely general to any kernel, inducing variable, and likelihood. Any code that is specific to a particular combination (e.g.~for efficiency) can still be specified, but external to the model and even the GPflow package itself (e.g.\ \citet{gpflow-vff-notebook}). This achieves our desiderata of efficiency, and extensibility.
In the next sections, we will describe the details of each component of the ELBO: covariances (\cref{sec:implementation:covariances}), \texttt{prior\_kl()} (\cref{sec:software:kl}), \texttt{conditional()} (\cref{sec:conditionals}), and \texttt{variational\allowbreak\_expectations()} (\cref{sec:implementation:likelihood}).

\subsubsection{Covariances $\Kuu$ \& $\Kuf$}
\label{sec:implementation:covariances}
Although not directly visible in \code{SVGP}, both \code{prior_kl()} and \code{conditional()} need to compute the covariances $\Kuu$ and $\Kuf$. Since their expressions (\cref{eq:id-covariances-def}) depend on both the type of the kernel and inducing variables, we also implement their interface with multiple dispatch (\cref{lst:multidispatch-covariance-example}).
\begin{lstlisting}[language=Python, style=mycodestyle,caption={Three definitions for interdomain covariances, using multiple dispatch for extensibility.},captionpos=b,label=lst:multidispatch-covariance-example]
@Kuu.register(InducingPoints, Kernel)
def Kuu_kernel_inducingpoints(inducing_variable: InducingPoints, kernel: Kernel, ...):
    ...


@Kuu.register(Multiscale, SquaredExponential)
def Kuu_sqexp_multiscale(inducing_variable: Multiscale, kernel: SquaredExponential, ...):
    ...


@Kuu.register(InducingPatches, Convolutional)
def Kuu_conv_patch(feat, kern, ...):
    ...
\end{lstlisting}

\subsubsection{Kullback--Leibler divergences}
\label{sec:software:kl}
To compute the KL divergence between the approximate posterior process $q(\mof(\cdot))$ and the prior process $p(\mof(\cdot))$, we only need to compute the finite-dimensional KL divergence at the inducing points \citep{matthews2016sparse}:
\begin{align}
\KL{q(\vu)}{p(\vu)} &= \Exp{q(\vu)}{\log q(\vu) - \log p(\vu)} \\
&=  -\tfrac{1}{2}\log|\mS|  -\tfrac{M}{2} +\tfrac{1}{2}\log |\Kuu|+\tfrac{1}{2} \Tr \big( \Kuu\inv (\mS + \vm \vm^\top)\big) \label{eq:kl-computation}
\end{align}
Due to the possible structure in $\Kuu$, we use multiple dispatch on the inducing variables and kernel. This gives a general interface for \code{prior_kl()} of:
\begin{lstlisting}[language=Python, style=mycodestyle,captionpos=b,label=lst:multidispatch-priorkl]
def prior_kl(inducing_variables, kernel, q_mu, q_sqrt, whiten=False):
    ...
\end{lstlisting}

For unstructured $\Kuu$, GPflow provides a default \code{gauss_kl()} function to compute \cref{eq:kl-computation}. In the formulation above, it requires the $\Kuu$ matrix, which is computed with the multiple dispatch function \code{Kuu}, together with the variational parameters $\vm$ and $\mS$. We parameterise $\mS$ through its lower-triangular Cholesky factorisation, i.e.~$\mS = \LS\LS\transpose$, which makes $\log\,\detbar{\mS} = 2\sum_{m=1}^M \log \left[\LS\right]_{mm}$, which is computationally cheaper.

We can alternatively compute the KL without direct access to $\Kuu$ by \emph{whitening} the inducing variables. In \code{SVGP} this is indicated by setting \code{whiten=True}. Instead of conditioning on $\vu$, we use a transformed version $\vv$ designed to have an identity covariance in the prior:
\begin{align}
    \vu = \Luu \vv\,, \qquad p(\vv) = \NormDist{\vzero,\eye}\,,
\end{align}
where $\Kuu = \Luu \Luu\transpose$. It is clear that this construction implies the correct distribution $p(\vu) = \mathcal N(0, \Kuu)$. If we represent the variational posterior over $\vv$ instead of $\vu$, then
\begin{align}
    \q{\vv} = \NormDist{\vm_{\vv}, \mS_{\vv}}\quad \implies \quad q(\vu) &= \NormDist{\Luu \vm_\vv, \Luu \mS_\vv \Luu\transpose}\,.
\end{align}
Since the prior over $\vv$ is now a centered standard Normal, this simplifies the KL divergence:
\begin{align}
\KL{q(\vu)}{p(\vu)} &=  \KL{q(\vv)}{p(\vv)}\\
&=-\tfrac{1}{2}\log| \mS_\vv|  -\tfrac{M}{2} +\tfrac{1}{2} \vm_\vv\transpose \vm_\vv +  \tfrac{1}{2} \Tr\big(\mS_\vv\big) .
\end{align}
Whitening can make optimisation or MCMC sampling more efficient \citep{hensman2015mcmc}. In the GPflow source code a \code{whiten} flag indicates whether we are passing ($\vm_\vv$, $\mS_\vv$) or ($\vm$, $\mS$).

\subsubsection{Conditionals}
\label{sec:conditionals}
The computation of the approximate posterior distribution $q(\mofp{X})$ is the most complex component to be specified. It is required for computing the expected log-likelihood (\cref{sec:implementation:likelihood}), as well as for making predictions. In its most complete form, $q(\mofp{X})$ is specified by covariances between every output for every input, giving $N\times P\times N\times P$ elements in total. Computing the expected log-likelihood never requires covariances between outputs for different inputs, and if the likelihood factorises over the output dimensions, only the $P$ marginal variances are needed. Covariances between outputs for different inputs are generally only needed for plotting. This gives four different predictive covariances that could be computed, determined by whether covariances are needed for different inputs (\code{full_cov=\{True|False\}}), and outputs (\code{full_output_cov=\{True|False\}}).

We bring all this functionality into the single function \code{conditional()}, which deals with computing $q(\mofx)$ with all four output covariance shapes. From the point of use in \code{SVGP}, as described in \cref{sec:gen-structure}, we need a fixed signature for \code{conditional()}, which we specify as:
\begin{lstlisting}[language=Python, style=mycodestyle]
def conditional(Xnew: Tensor, inducing_variable: InducingVariables, kernel: Kernel, f: Tensor, *, q_sqrt=None, full_cov=False, full_output_cov=False, white=False):
    """
    Parameters:
        Xnew: Locations at which to evaluate the approximate posterior GP
        inducing_variable: inducing variable of the approximate posterior, e.g., the inducing inputs Z.
        kernel: the model's kernel
        f: mean on which to condition
        q_sqrt: Cholesky factor of the covariance on which to condition
    """
\end{lstlisting}
The return values of \code{conditional()} are the mean and covariance of function values for the $N$ input points in \code{Xnew}. The return shape of the covariance depends on the value of \code{full_cov} and \code{full_output_cov}, summarised in \cref{tab:conditional-cov-output-shape}. We choose the layout of the returned covariances based on what is most convenient for follow-on operations. Since these are usually matrix multiplications or Cholesky decompositions, the tailing dimensions always contain the matrix of covariances that are requested.

\begin{table}[tb]
\centering
\begin{tabular}{llll}
\hline
\code{full_cov} & \code{full_output_cov} & Shape & Example use case  \\
\hline
\code{True}     & \code{True}  & $N\!\times\! P\times\! N\!\times\! P$ & Cholesky for sampling on $NP\!\times\! NP$.  \\
                & \code{False} & $P\!\times\! N\!\times\! N$         & Batch Cholesky for sampling.  \\
\code{False}    & \code{True}  & $N\!\times\! P\!\times\! P$         & Matmul for predictive density. \\
                & \code{False} & $N\!\times\! P$                     & Log-likelihood expectation. \\
\hline
\end{tabular}
    \caption{Shapes of the covariance returned by conditionals for different values of \code{full_cov} and \code{full_output_cov}. The returned mean always has shape $N\!\times\!P$.}
    \label{tab:conditional-cov-output-shape}
\end{table}

In principle, \code{conditional()} only needs to implement \cref{eq:qmox}. However, different combinations of priors and inducing variables induce different structure in $\Kuu$ and $\Kuf$, which can be exploited for computational efficiency. As described earlier, we will use multiple dispatch to allow the most efficient code path to be specified, while keeping a consistent interface. We dispatch on the specific \code{inducing_variable} and \code{kernel} that are passed to \code{conditional()}. In the following sections, we describe three different conditionals that are implemented in GPflow, each making a different assumption about $\Kuu$ and $\Kuf$. In \cref{sec:example-models}, we discuss various models (including the LMC from \cref{sec:example-lmc}) that can be implemented efficiently using these and other custom conditionals.

\subsubsection{Likelihood}
\label{sec:implementation:likelihood}
The expected log-likelihood is computed by \texttt{Likelihood.variational\allowbreak \_expectations\allowbreak (Fmu, Fvar, Y)}, which implements the integral
\begin{equation}
    \Exp{\mofxn}{\log p(\vy_n\given\mofxn} = \int \NormDist{\mofxn ; \vmu_n, \mSigma_n} \log p(\vy_n \given \mofxn) \calcd{\mofxn} \,.
\end{equation}
This function is separated from all other components as it depends only on the variational predictive distribution $q(\mofxn)$. This is computed by \code{conditional()}, and represented through its mean $\vmu_n \in \Reals^{P}$ (passed as \code{Fmu}) and covariance $\mSigma_n \in \Reals^{P \times P}$ (passed as \code{Fvar}). %

For certain likelihoods this expectation can be calculated in closed form. This is implemented by inheriting from the \code{Likelihood} base class, and implementing the closed-form expression in \texttt{Likelihood\allowbreak .variational\allowbreak \_expectations\allowbreak (Fmu, Fvar, Y)}. If this not possible, \GPflow can fall back to either Gauss-Hermite quadrature or Monte Carlo estimation. If $\mSigma_n$ is diagonal (i.e.~the outputs under the variational posterior are independent), or if the likelihood is independent over dimensions as $\log p(\vy_n\given \mofp{\vx_n}) = \sum_p \log p(y_{np}\given f_p(\vx_n))$, the multidimensional integral becomes equivalent to a sum over single dimensional integrals, which makes Gauss-Hermite or Monte Carlo estimation particularly effective. These can be used by inheriting from the classes \code{Likelihood} or \code{MonteCarloLikelihood}, respectively.

\section{Implementation of example methods in our framework}
\label{sec:example-models}
In this section, we wish to demonstrate the efficiency, modularity, and extensibility of our framework by discussing several example methods. We pay particular attention to demonstrating the use of multiple dispatch, and show how the appropriate \code{Kuu()}, \code{Kuf()}, and \code{conditional()} functions can be implemented.

\subsection{Single-output Gaussian processes}
\label{sec:single-output-conditional}
We begin with the most simple example of a single-output GP with an approximate posterior that induces no structure in $\Kuu$. \Cref{lst:api-single-output} sets up an example of such a model with an inducing point posterior, with \code{Kuu()} and \code{Kuf()} being defined in \cref{lst:multidispatch-covariance-example}. To complete the implementation we need to define a version of \code{conditional()} and register it using \code{@conditional.register(...)}. Since there is no structure in $\Kuu$, we need to define a version of conditional that simply implements \cref{eq:qx} in accordance with the shapes in \cref{tab:conditional-cov-output-shape}. We register the function using the following code:
\begin{lstlisting}[language=Python, style=mycodestyle]
@conditional.register(object, InducingVariables, Kernel, object)
@name_scope("conditional")
def singleoutput_conditional(Xnew, inducing_variable, kernel, f, *, full_cov=False, full_output_cov=False, q_sqrt=None, white=False):
    Kmm = Kuu(inducing_variable, kernel, jitter=settings.jitter)  # [M, M]
    Kmn = Kuf(inducing_variable, kernel, Xnew)  # [M, N]
    Knn = kernel.K(Xnew) if full_cov else kernel.Kdiag(Xnew)  # [N, N] or [N] 

    fmean, fvar = base_conditional(Kmn, Kmm, Knn, f, full_cov=full_cov,
                                   q_sqrt=q_sqrt, white=white)  # [N, 1],  [1, N,  N] or [N, 1]
    return fmean, _expand_independent_outputs(fvar, full_cov, full_output_cov)
\end{lstlisting}
We register the function as a conditional with the dispatcher for general \code{Kernel} and \texttt{Inducing\allowbreak Variable} objects, using \code{@conditional.register(...)}. This will make this code path the default for all kernels and inducing variables that inherit from the base classes \code{Kernel} and \texttt{Inducing\allowbreak Variable}, unless more specialised implementations are provided.
The main work of implementing \cref{eq:qx} is done in \texttt{base\_conditional()}. We pass in $\Kuu$ and $\Kuf$ as matrices, after they have been evaluated using the multiple dispatched functions \code{Kuu()} and \code{Kuf()}. To keep \code{base_conditional()} simple, it does not implement \texttt{full\_output\_cov=True}. To ensure that the returned shapes variances are consistent with other multioutput conditionals, we use the helper function \texttt{\_expand\allowbreak \_independent\allowbreak \_outputs()}.

\subsection{Multiscale interdomain inducing variables}
In \cref{sec:interdomain} we discussed the use of integral transformations of a function $\ff$ as inducing variables. 
\citet{lazaro2009inter} proposed using $w_m(\vx)$ being a Gaussian, or a Gaussian-windowed sinusoid. When combined with the \code{SquaredExponential} kernel, the integrals for the required covariances (\cref{eq:transdomain-cov,eq:interdomain-cov}) have closed form expressions. We included the Gaussian integral transformation as the \code{Multiscale} inducing variable in \GPflow. While these inducing variables do not provide special structure in $\Kuu$ and $\Kuf$, they do allow the posterior to represent longer-range correlations with fewer inducing variables.
In terms of code, this method can be implemented by simply specifying the appropriate covariances:
\begin{lstlisting}[language=Python, style=mycodestyle]
class Multiscale(InducingPoints):
    ...

@Kuf.register(Multiscale, kernels.SquaredExponential, object)
def Kuf(inducing_variable, kernel, Xnew):
    ...

@Kuu.register(Multiscale, kernels.SquaredExponential)
def Kuu(inducing_variable, kernel, *, jitter=0.0):
    ...
\end{lstlisting}
Using the \code{@method.register(...)} line, we ensure that this code is specific to the combination of the \code{Multiscale} inducing variable, and the \texttt{Squared\allowbreak Exponential} kernel. Since \code{Multiscale} and \texttt{Squared\allowbreak Exponential} inherit from \code{InducingPoints} and \code{Kernel} respectively, the same version of \code{conditional(...)} will be called as in the previous section.

\subsection{Image convolutional Gaussian processes}
Similarly to \code{Multiscale} inducing variables, the inducing patches for the convolutional Gaussian process \citep{vdw2017convgp} can be implemented using by only specifying \code{Kuu()} and \code{Kuf()}, while relying on the default conditional. We define \code{InducingPatch} for specifying inducing variables in $g(\cdot)$ (see \cref{sec:convgp}), and define multiple dispatch \code{Kuu} and \code{Kuf} functions for use with the \code{Convolutional} kernel:
\begin{lstlisting}[language=Python, style=mycodestyle]
@Kuf.register(InducingPatch, kernels.Convolutional, object)
def Kuf(inducing_variable, kernel, Xnew):
    ...

@Kuu.register(InducingPatch, kernels.Convolutional)
def Kuu(inducing_variable, kernel, ...):
    ...
\end{lstlisting}
Image-to-image convolutional Gaussian processes \citep{dutordoir2019tick} also fit within this framework, but require custom conditionals to implement the covariances between the different outputs.

\subsection{Variational Fourier features}
\citet{hensman2017variational} define inducing variables in the spectral domain of the process which lead to periodic functions for $\Kuf$. More importantly, this careful choice leads to a ``low-rank plus diagonal'' form of $\Kuu$, i.e.
\begin{equation}
\Kuu = \diag(\valpha) + \vbeta\vbeta\transpose \,,
\end{equation}
where $\valpha \in \Reals^M$ and $\vbeta \in \Reals^{M\times d}$, where $d-1$ is the mean-squared differentiability of the functions in the prior.

This method naturally fits into \GPflow's interdomain and multiple dispatch framework. We can define a new type of feature, \code{FourierFeatures}, which inherits from \code{InducingVariable}. It would be mathematically possible and an immediate fit into the existing framework if we implement a new \code{Kuu()} method which computes the full (non-structured) $\Kuu$, and register it within the multiple-dispatch framework.

This approach, however, is sub-optimal and would not exploit any of the known structure (diagonal $+$ low rank) of $\Kuu$. To take advantage of this, we need to route the model through more specialised code for both the KL divergence and the conditional.
We let the covariance method for $\Kuu$ return an object that captures the structure of $\diag(\valpha) + \vbeta\vbeta\transpose$ (in this case, a TensorFlow \code{LinearOperator}). We then register a specialised \code{conditional} method that can deal with subsequent computations in the computationally most efficient way given its access to $\diag(\alpha)$ and $\beta$.

The covariances can be implemented closely following the equations in \cite{hensman2017variational}, e.g.\ for the Mat\'ern 1/2 kernel:
Note that different orders of Mat\'ern kernels have different structures in their $\Kuu$ matrix, so we have to separately register one implementation for each combination of \code{FourierFeatures} inducing variable and one of the Mat\'ern kernels. Similarly, we need to register separate implementations for $\Kuf$. The conditional, however, only needs to know that the solve operation is efficient.
A sample implementation is given in the \citet{gpflow-vff-notebook}.

\subsection{Linear Model of Coregionalization}
\label{sec:lmc-implementation}

\subsubsection{Multioutput inducing points}
\label{sec:lmc-implementation-inducingpoints}
When moving to specifying approximate posteriors for multioutput kernels, we have to deal with all the possible covariance structures described in \cref{tab:conditional-cov-output-shape}. We start with the inducing point method for multioutput GPs, which was described in \cref{sec:example-lmc-inducing-points} and applied to the LMC. This method takes all $\Mout = MP$ output values from $M$ inputs as inducing variables. The single-output conditional (\cref{sec:single-output-conditional}) does not compute the required covariances for \code{full_output_cov=True}, so we will need define a new \code{conditional()}.

To implement the predictive equation (\cref{eq:qmox}), \code{Kuu()} needs to compute all $M\!\times\!P\!\times\!M\!\times\!P$ covariances, while \code{Kuf()} needs to compute the $M\!\times\!P\!\times\!N\!\times\!P$ covariances between $\vu$ and the predictions. We let these functions simply compute and return the covariances defined in the kernel. Since this implementation of the \code{InducingPoints} inducing variable is general to all multioutput kernels, we register the implementation for all kernels that inherit from the multioutput kernel base class \code{MultioutputKernel}:
\begin{lstlisting}[language=Python, style=mycodestyle]
@Kuu.register(InducingPoints, Multioutputkernel)
def Kuu(inducing_variable, kernel, *, ...):
    Kmm = kernel.K(inducing_variable.Z, full_output_cov=True)  # [M, P, M, P]
    ...
    return Kmm
    
@Kuf.register(InducingPoints, MultioutputKernel, object)
def Kuf(inducing_variable, kernel, Xnew):
    return kernel.K(inducing_variable.Z, Xnew, full_output_cov=True)  #  [M, P, N, P]
\end{lstlisting}

We now require an implementation of \code{conditional()} that can use these differently shaped covariances. When \code{full_cov} and \code{full_output_cov} are equal, i.e.~we only care about the marginals, or we want the complete joint over both the inputs and outputs, we can rely on the \code{base_conditional()} that we used earlier after a simple reshape. For the other cases, we implement \code{fully_correlated_conditional()}. As with the implementation of the covariances, we define this \code{conditional()} to be the default for all multioutput kernels when \code{InducingPoints} are used. For this example, we can assume that \code{P} is equal to \code{L}.
\begin{lstlisting}[language=Python, style=mycodestyle]
@conditional.register(object, InducingPoints, MultioutputKernel, object)
def _conditional(Xnew, inducing_variable, kernel, f, *, full_cov=False, full_output_cov=False, q_sqrt=None, white=False):
    """
    ...
    :param f: variational mean, [ML, 1]
    :param q_sqrt: standard-deviations or cholesky, [ML, 1]  or  [1, ML, ML]
    """
    Kmm, Kmn = ...  # [M, L, M, L] ,  [M, L, N, P]
    if full_cov:
        Knn = kernel.K(Xnew, full_output_cov=full_output_cov)  # [N, P, N, P] (full_output_cov = True) or [P, N, N]
    else:
        Knn = kernel.Kdiag(Xnew, full_output_cov=full_output_cov)  # [N, P, P] (full_output_cov = True) or [N, P]

    M, L, N, P = [tf.shape(Kmn)[i] for i in range(Kmn.shape.ndims)]
    Kmm = tf.reshape(Kmm, (M * L, M * L))

    if full_cov == full_output_cov:
        Kmn = tf.reshape(Kmn, (M * L, N * P))
        Knn = tf.reshape(Knn, (N * P, N * P)) if full_cov else tf.reshape(Knn, (N * P,))
        fmean, fvar = base_conditional(Kmn, Kmm, Knn, f, full_cov=full_cov, q_sqrt=q_sqrt, white=white)  # NP x 1, 1 x NK(x NK)
        fmean = tf.reshape(fmean, (N, P))
        fvar = tf.reshape(fvar, (N, P, N, P) if full_cov else (N, P))
    else:
        Kmn = tf.reshape(Kmn, (M * L, N, P))
        fmean, fvar = fully_correlated_conditional(
            Kmn, Kmm, Knn, f, full_cov=full_cov,
            full_output_cov=full_output_cov, q_sqrt=q_sqrt, 
        )
    return fmean, fvar
\end{lstlisting}

In order to use this code with the LMC, all we have to do is define the covariance of the prior in an object that inherits from the multioutput base class \code{MultioutputKernel}:
\begin{lstlisting}[language=Python, style=mycodestyle]
class LinearCoregionalization(IndependentLatent, Combination):
    ...
    def K(self, X, Y=None, full_output_cov=True, presliced=False):
        ...
        if full_output_cov:  # Include covariances between outputs
            ...
            return ...  # [N, P, N, P]
        else:  # Marginals over outputs
            return ...  # [P, N, N]

    def Kdiag(self, X, full_output_cov=True):
        ...  # Do not return covariances between different inputs
        if full_output_cov:  # Include covariances between outputs
            ...
            return ...  # [N, P, P]
        else:  # Marginals over outputs
            return ...  # [N, P]
\end{lstlisting}

\subsubsection{Latent inducing points}
In \cref{sec:example-lmc-latent-inducing-points}, we discussed that placing the inducing variables in the process $\mog(\cdot)$ reduced the computational cost of the method.

The previous implementation focused on generality. Here we want to take advantage of additional structure. The \code{IndependentLatent} kernel allows inference with inducing variables \code{Fallback \{Shared|Separate\}InducingVariables}. These inducing variables give a block-diagonal structure in $\Kuu$, which is represented as a $L\times M\times M$ array (\cref{sec:example-lmc-latent-inducing-points}). To take advantage of this we define a custom conditional for these combinations of inducing variables and kernels. The exact details are abstracted away in \code{independent_interdomain_conditional(*)} which takes advantage of the block-diagonal structure.
\begin{lstlisting}[language=Python, style=mycodestyle]
@conditional.register(object, (FallbackSharedIndependentInducingVariables, FallbackSeparateIndependentInducingVariables), LatentIndependent, object)
@name_scope("conditional")
def _conditional(Xnew, feat, kern, f, *, full_cov=False, full_output_cov=False, q_sqrt=None, white=False):
    Kmm = Kuu(feat, kern, jitter=settings.numerics.jitter_level)  # [L, M, M]
    Kmn = Kuf(feat, kern, Xnew)  # [M, L, N, P]
    if full_cov:
        Knn = kernel.K(Xnew, full_output_cov=full_output_cov)  # [N, P, N, P] (full_output_cov = True) or [P, N, N]
    else:
        Knn = kernel.Kdiag(Xnew, full_output_cov=full_output_cov)  # [N, P, P] (full_output_cov = True) or [N, P]

    return independent_interdomain_conditional(
        Kmn, Kmm, Knn, f, q_sqrt=q_sqrt, full_cov=full_cov, full_output_cov=full_output_cov, white=white)
\end{lstlisting}

The previous implementation of LMC using the default \code{MultioutputKernel} and \code{InducingPoints} leads to a scaling of $\BigO\left(P^3M^3 + NP^3M^2\right)$. In this section, using the specialised codepath gives $\BigO\left(LM^3 + NPLM^2\right)$ (see \cref{sec:example-lmc-computational-complexity}).

\section{Uncertain inputs}

\subsection{Background}

So far we have discussed a set of models for which all inputs $\{\vx_n\}_{n=1}^N$ are known and fixed variables: they form the features of the model. This is the typical setting for most supervised learning algorithms.
In certain applications, however, the GP models have to deal with inputs which are \emph{uncertain}, and typically described by a distribution $\{p(\vx_n)\}_{n=1}^N$, rather than a single point.
A well-known example of this type of model is the GP-LVM \citep{lawrence2004gplvm}, where we are interested in learning the joint posterior of two variables: the distribution of the input locations $\{q_n(\vx_n)\}_{n=1}^N$ of the GP function and the GP mapping $f(\cdot)$ itself. 
More recent models such as the deep Gaussian process (DGP) \citep{damianou2013dgp} and the conditional density estimation model \citep{dutordoir2018cde} also require dealing with distributional or uncertain inputs.

Uncertain inputs are omnipresent in GP models and need to be accounted for in our software tool. In the following sections we discuss the optimisation bounds of the GP-LVM and DGP models, and show how they are implemented in \GPflow. The focus of this section is to show that even when constructing these more complex models, we can still follow our desiderata of efficiency, modularity and extensibility. This leads to a software package that can be used to \emph{extend} the functionality or modelling assumptions of DGPs by re-implementing only certain bits of the computational graph while staying inside the framework.
\subsection{Examples}

\subsubsection{Latent Variable Model}

The GP-LVM \citep{lawrence2004gplvm} is an unsupervised algorithm which learns for each of the given points in the dataset $\{\vy_n\}_{n=1}^N$ an associated input distribution $\{q_n(\vx_n)\}_{n=1}^N$, together with a GP mapping, so that $\vy_n = f(\vx_n) + \text{noise}$. This is different from the typical regression or classification settings, where both the input and outputs are known and the only quantity to infer is the underlying mapping.

Deriving the bound for the GP-LVM is similar as in \cref{sec:variational-inference-for-GPs}: the approximate GP is defined by a set of inducing variables, which are learned by minimising the KL between the true and the approximate posterior. To learn the approximate posterior of the inputs in the GP-LVM we follow the same procedure, and define an approximate posterior over the uncertain inputs $\{q_n(\vx_n)\}_{n=1}^N$. Following Bayes' rule, as in \cref{sec:variational-inference-for-GPs} we obtain the lower bound
\begin{equation}
\lb_\text{\textsc{LVM}} = \sum_{n=1}^N \Exp{q(\vx_n)}{\Exp{q(\mofxn)}{\log p(\vy_{n}\given \mofxn)}} - \sum_{n=1}^N \KL{q_n(\vx_n)}{p(\vx_n)} - \KL{q(\vu)}{p(\vu)}\,. \label{eq:elbo-uncertain-inputs}
\end{equation}
Most of the elements of this bound are already discussed in \cref{sec:implementation-svgp}, and stay unchanged in the case of uncertain inputs. However, the additional unknown variables and associated approximate posterior introduce two new components to the bound.

The first component is a KL term (for each datapoint) between the prior and approximate posterior of the uncertain inputs $\vx_n$. As with the KL term on the inducing variables, given reasonable choices for the form of distributions this quantity can be computed in closed form. 

The second component is an additional expectation of the data-fit term with respect to $q(\vx)$.
In general, solving this expectation analytically is not possible (as is the case for the inner expectation over $q(\mofxn)$ for most likelihoods). To deal with this we resort to Monte Carlo and sample $\widetilde{\vx_n} \sim q(\vx_n)$ to estimate the quantity.
Using re-parameterised samples \citep{kingma2013auto, rezende2014dglm} from the approximate posterior distribution allows us to get unbiased gradient estimates of the ELBO and optimise for the parameters of $q(\vx)$, simultaneously with the other parameters as before. 

For certain kernels and forms of $q(\vx_n)$ the expectation can be computed analytically by making use of the so-called kernel expectations \citep{damianou2016uncertain}. While this approach adds less stochasticity to the bound, it increases the computational costs and limits the set of kernels that can be used.

\subsubsection{Deep Gaussian processes}
A deep Gaussian process consists of multiple GPs which are hierarchically chained to form a compositional function, $\moff = \moff_L \circ \moff_{L-1} \circ \,\ldots\,\circ \moff_1$.
At each level the input of the current function is given by the output of the previous one. The uncertain inputs in a DGP arise from the fact that the hidden state that is used as the input to the next GP is given by the output of a GP, which is itself a distribution.

Imposing an independent GP prior over each of the components $p(\moff_\ell) = \GP(0, k_\ell(\cdot, \cdot))$ and hidden states $\vh_{\ell} = \moff_\ell(\vh_{\ell-1})$, the joint density of a DGP can be written as
\begin{gather}
    p(\vy, \moff_1(\cdot), \ldots, \moff_L(\cdot), \vh_1, \ldots, \vh_{L-1} \given X) = \prod_{n=1}^N p(\vy_n \given \vh_{n, L}) \prod_{\ell=1}^L p(\moff_\ell(\cdot))\,p(\vh_{n, \ell} \given \moff_\ell(\cdot), \vh_{n, \ell-1}),
\end{gather}
where we define $\vh_{n, 0} := \vx_n$. The form of $p(\vh_{n, \ell} \given \moff_\ell(\cdot), \vh_{n, \ell-1})$ leads to different types of DGPs. The original DGP formulation added noise between the layers, setting $p(\vh_{n, \ell} \given \moff_\ell(\cdot), \vh_{n, \ell-1})$ to $\NormDist{\vh_{n, \ell} \given \moff_\ell(\vh_{n, \ell-1}), \sigma_\ell^2}$ \citep{damianou2013dgp}. 
In \citet{salimbeni2017doubly} a noise-less transition is assumed, leading to $p(\vh_{n, \ell} \given \moff_\ell(\cdot), \vh_{n, \ell-1}) = \delta\left({\vh_{n, \ell} - \moff_\ell(\vh_{n, \ell-1})}\right)$. 

As was the case for shallow GP models (see \cref{sec:single-approx-inference} and \cref{sec:multi-approx-inference}), we construct an approximate posterior GP by conditioning the prior GP on a set of inducing variables. The inducing variables $\vu_\ell$, for which we have a set for each layer, are then integrated out with respect to their posterior $q(\vm_\ell, \mS_\ell)$, leading to an approximate posterior of the form $\NormDist{\mu_\ell(\vh_{\ell -1 }), \Sigma_\ell(\vh_{\ell - 1}, \vh_{\ell - 1})}$ where the mean and the variance are given by
\begin{align}
    \mu_\ell(\vh_{\ell - 1}) &= \K_{\vu_\ell \moff_\ell(\vh_{\ell-1})} ^ \top \K_{\vu_\ell \vu_\ell}\inv \vm_\ell,\\
    \Sigma_\ell(\vh_{\ell - 1}, \vh_{\ell - 1}) &= k_\ell(\vh_{\ell-1}, \vh_{\ell-1}) - \K_{\vu_\ell \moff_\ell(\vh_{\ell-1})}^\top \K_{\vu_\ell \vu_\ell}\inv \left(\K_{\vu_\ell \vu_\ell} - \mS_\ell \right) \K_{\vu_\ell \vu_\ell}\inv \K_{\vu_\ell \moff_\ell(\vh_{\ell-1})}.\label{eq:qdeep}
\end{align}
The crucial difference between this approximate posterior and \cref{eq:qmox} is that here the inputs to the mean and variance are the outputs of the previous layer $\vh_{\ell - 1}$, and thus \emph{uncertain} variables. %
As was the case in the GP-LVM model, integrating out the inputs to account for their uncertainty is not possible due to the non-linear mapping of the inputs in the kernel. However, when appropriately propagating the uncertainty through the layers of the DGP, it is possible to compute a sample at each layer:
\begin{equation}
    \widetilde{\vh}_\ell = \mu_\ell(\widetilde{\vh}_{\ell-1}) + {\Sigma_\ell^{\frac12}(\widetilde{\vh}_{\ell-1}, \widetilde{\vh}_{\ell-1})}\,\epsilon,\quad\text{where}\quad\,\epsilon \sim \NormDist{0, \eye} .
    \label{eq:sample-layer}
\end{equation}

Using the approximate posterior for every GP in the composition, we can lower-bound the log marginal likelihood to get the ELBO for a DGP:
\begin{equation}
\lb_\text{\textsc{DGP}} = \sum_{n=1}^N \Exp{q(\vh_L)}{\log p(\vy_{n}\given \vh_L)} - \sum_{\ell=1}^L \KL{q(\vu_\ell)}{p(\vu_\ell)}\,. \label{eq:elbo-deepgp}
\end{equation}
In this bound, compared to the shallow GP case (\cref{eq:elbo-gp}), we are dealing with a variational expectation over the output of the last layer $q(\vh_L)$ and a sum over the KL terms of the inducing variables of every layer.
Evaluating the KLs in this bound is straightforward given the Gaussianity of both prior and approximate posterior of $\vu_\ell$. The required computation is identical to the one required in the shallow case. The expectation over the final layer is approximated using Monte Carlo. Following the procedure of \cref{eq:sample-layer}, we obtain a sample from $q(\vh_L)$ which is used to get an unbiased estimate of the variational expectation, as described in \citet{salimbeni2017doubly}.

\subsection{Implementation}
Implementing bounds for models with uncertain inputs, e.g.\ the GP-LVM (\cref{eq:elbo-uncertain-inputs}) and DGP (\cref{eq:elbo-deepgp}) is not significantly harder, as most of the required components are identical to the building blocks of the SVGP model discussed in \cref{sec:implementation-svgp}.

The most notable difference between the standard SVGP model and models with uncertain inputs is that for the SVGP, we compute the mean and variance of $q(\mofxn)$ at known locations $\{\vx_n\}_{n=1}^N$, whereas for models with uncertain inputs, we evaluate the posterior GP at samples from the source of uncertainty. This can be the distribution of the inputs $q(\vx_n)$ in the case of the GP-LVM or the output distribution of the previous layer $q(\vh_\ell)$ for DGPs.

To accommodate this use-case, GPflow provides a method called \texttt{sample\_conditional}, whose role is to compute a sample from the approximate posterior $q(\mofx)$. The method has the same signature as \texttt{conditional} and in most cases will use \texttt{conditional} to compute the mean $\vmu$ and covariance $\mSigma$ of the approximate posterior, so that it can construct a sample $\vmu + \mSigma^{\frac12}\,\epsilon$, with $\epsilon \sim \NormDist{0, \eye}$.
However, for certain types of kernels and inducing variables it is possible to compute a valid sample from the posterior in a more efficient way, which is why \texttt{sample\_conditional} itself makes use of multiple dispatch. As was the case for  \texttt{conditional} (see \cref{sec:conditionals} and \cref{sec:multipedispatch}), thanks to multiple dispatch, specialised code can be executed depending on the types of the arguments to the method in order to compute the sample in the most efficient way.

\paragraph{Modularity and extensibility}
The \texttt{sample\_conditional} method computes a sample from the GP's approximate posterior, which can then be propagated through the layers of a DGP to obtain a valid sample of the composite function. A user can alter the behaviour of a layer in a DGP by defining a new kernel and inducing variable, together with the corresponding \texttt{sample\_conditional}. This makes GPflow easy to extend and modular, as evidenced by e.g.~\citet{blomqvist2018dcgp,dutordoir2019tick} who implemented deep convolutional (multi-output) GPs following this approach, and \citet{salimbeni2019deep} who implemented a DGP with latent variable layers.

\section{Conclusion}

In this work we presented the current state of Gaussian process models. 
We reviewed current modelling assumptions such as multi-output GPs and discussed methods for efficient inference: sparse variational inference and inter-domain inducing variables.
We presented this in a unified mathematical framework, starting from single-output models up to multi-output inter-domain GPs.
Throughout the text it becomes clear that efficient inference in GP models relies strongly on choosing useful inducing variables in a way that can take computational advantage of independence properties of the prior. To this end, we developed a software framework in \GPflow that allows for the flexible specification of both multi-output priors and inducing variables.

\subsubsection*{Acknowledgements}
Many thanks to Felix Leibfried for repeatedly reading drafts and providing excellent feedback. Also thanks to Kaspar M{\"a}rtens for providing feedback on early versions of the manuscript. Thanks to the TensorFlow Probability team and the GPflow community for their work and support.

\printbibliography

\end{document}